2026-05-09

# Probability Bracket Notation: Multivariable Systems and Static Bayesian Networks

Xing M. Wang[1]

## Abstract

We extend Probability Bracket Notation (PBN), inspired by the Dirac notation in quantum mechanics, to multivariable probability systems and static Bayesian networks (BNs). By defining probability distributions and conditional expectations in a unified, basis-independent algebraic form, PBN provides a systematic way to represent and manipulate dependencies among random variables. Using the well-known Student BN as an illustrative probabilistic graphical model, we demonstrate prediction, bottom-up and top-down inference, and expectation calculations within the PBN framework. We show that, for a large N-node binary BN, after a one-time preprocessing, inference along a d-separable chain with $k$ intermediate nodes requires $O(k{\cdot}2^k)$ operations, compared to $O(N{\cdot}2^N)$ for direct computation from the full joint distribution. We further extend PBN to networks with continuous variables, including linear Gaussian models, and introduce a hybrid Healthcare BN that combines discrete and continuous variables. In this model, discrete-display nodes serve as proxies for continuous parents, enabling user-specific predictions. Overall, PBN provides an operator-based framework that unifies representation and computation, with potential applications in education, data analytics, and machine learning.

## 1. Introduction

Inspired by Dirac's notation, the Probability Bracket Notation (PBN) was introduced for probability modeling, in which we consider systems with a single variable or mutually independent variables [1]. We now apply PBN to multivariable systems within static Bayesian networks (BNs) [2-6]. For simplicity, we will not discuss dynamic Bayesian networks with PBN here; that topic will be addressed in our ongoing work [7].

In the **next section,** we introduce various probability distributions (PDs), including joint, marginal, intermediate, and conditional probability distributions (JPD, MPD, IPD, and CPD) [2]. We explore in more detail the relationships among the different PDs of the subnets of the well-known Student BN [3-6] using PBN. In **Section 3**, we analyze the dependency, local independence, and reasoning capabilities of the entire Student BN, including a set of reasoning exercises, and demonstrate how to use PBN to manipulate various probability expressions within the network. In **§3.4**, we investigate the power and the rule of inserting P-identities by introducing *d-separable chains*. We show that, after preparation, in a large $N$-node binary Bayesian network, PBN can reduce the number of calculations from $O(N{\cdot}2^N)$ using full JPD to $O(k{\cdot}2^k)$ for a d-separable inference chain with $k$ intermediate nodes. The two-phase procedure is particularly effective for d-separable chains in *extended polytrees* with pendant subnets,

[1] Sherman Visual Lab, Sunnyvale, CA 94085, USA, xmwang@shermanlab.com; ORCID:0000-0001-8673-925X

highlighting the role of network structure in reducing computational complexity (see Fig. 8). **Section 4** examines BNs with both continuous and discrete random variables [9-15]. After reviewing linear Gaussian networks, we propose a Healthcare BN that accepts user-specific inputs and produces customized predictions using discrete-display (DD) nodes (see Fig. 11). **Appendix A** provides details of the two-phase inference procedure.

## 2. Probability Bracket Notation and Multivariable Systems

In PBN, the random variables and their corresponding *P*-identities are all treated as operators, which can be inserted into suitable *P*-brackets [1]. In this section, we will examine multivariate probability spaces in PBN, with a particular focus on P-identities. For definitions of *P*-bra, *P*-ket, *P*-bracket, *P*-identity, and *P*-basis, we refer readers to Eqs. (8-22) of [1]. We will use the subnets of the Student BN [3-5] as examples. Let's begin with the system of two variables.

### 2.1. Two–Variable Systems and the Student I-S Example

Assume two discrete random variables, *X* and *Y*. Their sample spaces (domains) $\Omega_X$ and $\Omega_Y$ form a joint sample space $\Omega = \{\Omega_X, \Omega_Y\}$, having the following *P*-basis:

$$\begin{aligned}
&\text{For } \forall\{x, y\} \in \Omega: \ X \mid x, y) = x \mid x, y), \quad Y \mid x, y) = y \mid x, y), \quad P(\Omega \mid x, y) = 1 \\
&\text{Orthonormality: } P(x, y \mid x', y') = \delta_{xx'}\delta_{yy'}; \\
&\text{Completeness: } \quad \sum\nolimits_{x,y\in\Omega} \mid x, y) P(x, y \mid \ = I_{X,Y}
\end{aligned} \tag{1}$$

**2.1.1. The Definitions and Properties of Two-Variable PDs:** The (full) joint probability distribution (JPD) [2] is defined by:

$$\begin{aligned}
&P(x, y) = P_{X,Y}(x, y) = P(x, y \mid \Omega) = P(X = x, Y = y \mid \Omega), \quad \text{for } \forall\{x, y\} \in \Omega \\
&\{x, y\} \in \Omega \Leftrightarrow \{(x \in \Omega_X) \wedge (y \in \Omega_Y)\}
\end{aligned} \tag{2}$$

The full JPD has the following property:

$$P(x, y) \geq 0, \quad \text{if } \{x, y\} \in \Omega; \quad P(x, y) = 0, \quad \text{if } \{x, y\} \notin \Omega \tag{3}$$

Its normalization can be obtained by inserting the *P*-identity as follows:

$$1 = P(\Omega \mid \Omega) = P(\Omega \mid I_{X,Y} \mid \Omega) = \sum\nolimits_{x,y\in\Omega} P(\Omega \mid x, y) P(x, y \mid \Omega) = \sum\nolimits_{x,y\in\Omega} P(x, y) \tag{4}$$

The expectation value of a well-defined real function $F(X,Y)$ is given by:

$$E[F(X,Y)] \equiv P(\Omega \mid F(X,Y) \mid \Omega) = P(\Omega \mid F(X,Y) I_{X,Y} \mid \Omega) = \sum_{x,y\in\Omega} P(x, y) F(x, y) \tag{5}$$

The conditional expectation of $F(X,Y)$ given a subset $H \subset \Omega$ can be easily derived:

$$E[F(X,Y)\,|\,H] \equiv P(\Omega\,|\,F(X,Y)\,|\,H) = P(\Omega\,|\,F(X,Y)I_{X,Y}\,|\,H)$$
$$= \sum\nolimits_{\{x,y\}\in H} F(x,y)P(x,y\,|\,H) = \sum\nolimits_{\{x,y\}\in H} F(x,y)P(x,y)/P(H) \quad (6)$$

From the JPD in Eq. (2), we obtain the MPDs [2] for $x$ and $y$:

$$P(x) = \sum\nolimits_y P(x,y\,|\,\Omega) = \sum\nolimits_y P(x,y),\ \ P(y) = \sum\nolimits_x P(x,y\,|\,\Omega) = \sum\nolimits_x P(x,y) \quad (7)$$

In addition, we have the CPD [2] of $X$ given $Y$ and the CPD of $Y$ given $X$:

$$P(x\,|\,y) \equiv P(X=x\,|\,Y=y) = P_{X|Y}(x\,|\,y), \quad P(y\,|\,x) = P_{Y|X}(y\,|\,x) \quad (8)$$

Please do not confuse Eq. (8) with $P(x \mid x')$, where both event $x$ and evidence $x'$ are the observed values of the same random variable ($X$).

**Definition 2.1.1**: Event $X = x$ is independent of event $Y = y$ in a distribution $P$, denoted $P \models x \perp y$, if $P(x\,|\,y) = P(x)$ or if $P(y) = 0$ (§2.1.4.1 of [3]). (9)

**Definition 2.1.2**: variables $X$ and $Y$ are independent in a distribution $P$, denoted $P \models X \perp Y$, if $P \models x \perp y$ for $\forall x, y \in \Omega$. (10)

**Proposition 2.1.1**: $P \models X \perp Y$ if and only if $P(x,y) \equiv P(x \wedge y) = P(x)P(y)$. (11)
Its proof can be found in §2.1.4.3 of Ref [3].

Graphically, it is represented in Fig. 1:

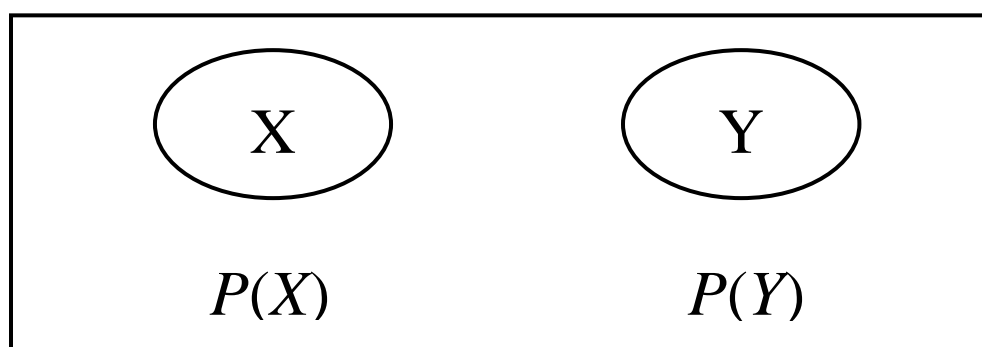

**Fig. 1**: $P \models X \perp Y$

The joint distribution is the product of the distributions of each node in Fig. 1:

$$P(X,Y) = P(X)\,P(Y) \quad (12)$$

In many cases, they are dependent, and the joint PD is unknown. We need to derive it. Suppose we are given $P(x\,|\,y)$ and $P(y)$, then we can use Bayes' chain rule to get it:

$$P(x,y) = P(x\,|\,y)\,P(y) \quad (13)$$

Then, we can derive the MPD for $X$:

$$P(x\,|\,\Omega) = P(x) = \sum\nolimits_y P(x,y) = \sum\nolimits_y P(x\,|\,y)\,P(y) = \sum\nolimits_y P(x\,|\,y)\,P(y\,|\,\Omega) \quad (14)$$

Because Eq. (14) is valid for $\forall x \in \Omega$, we have recovered the *P*-identity for the corresponding marginal probability distribution of $(\Omega_Y, Y, P)$:

$$P(x|\Omega) = \sum\nolimits_y P(x|\,y)P(y|\Omega) = P(x|\left\{\sum\nolimits_y |\,y)P(y|\right\}|\Omega) = P(x|I_Y|\Omega) \tag{15}$$

Or: $$P(x) = P_{X|\Omega}(x|\Omega) = P(x|I_Y|\Omega), \quad I_Y = \sum\nolimits_y |\,y)P(y| \tag{16}$$

Similarly, we get the following *P*-identity for *X*:

$$P(y) = P_{Y|\Omega}(y|\Omega) = P(y|I_X|\Omega), \quad I_X = \sum\nolimits_x |\,x)P(x| \tag{17}$$

Using the *P*-identities, we obtain the normalizations of Conditional PDs (CPs):

$$\begin{aligned} 1 &= P(\Omega|\,x) = P(\Omega|I_Y|\,x) = \sum\nolimits_y P(\Omega|\,y)P(y|\,x) = \sum\nolimits_y P(y|\,x) \\ 1 &= P(\Omega|\,y) = P(\Omega|I_X|\,y) = \sum\nolimits_x P(\Omega|\,x)P(x|\,x) = \sum\nolimits_x P(x|\,y) \end{aligned} \tag{18}$$

The normalization of the full JPD can also be found by inserting *P*-identities:

$$\begin{aligned} 1 &= P(\Omega|\Omega) = P(\Omega|I_X|\Omega) = \sum\nolimits_{x\in\Omega} P(x|\Omega) = \sum\nolimits_{x\in\Omega} P(x|I_Y|\Omega) \\ &= \sum\nolimits_{x,y\in\Omega} P(x|\,y)P(y|\Omega) \underset{(13)}{=} \sum\nolimits_{x,y\in\Omega} P(x,y) = P(\Omega|I_{X,Y}|\Omega) \end{aligned} \tag{19}$$

**2.1.2. The *I*→*S* Example of the Student Bayesian Network:** Consider an example from the Student BN [3], containing variables *I* (student Interagency scores) and *S* (student SAT scores). We assume *S* depends only on *I*. It is represented graphically in Fig. 2.

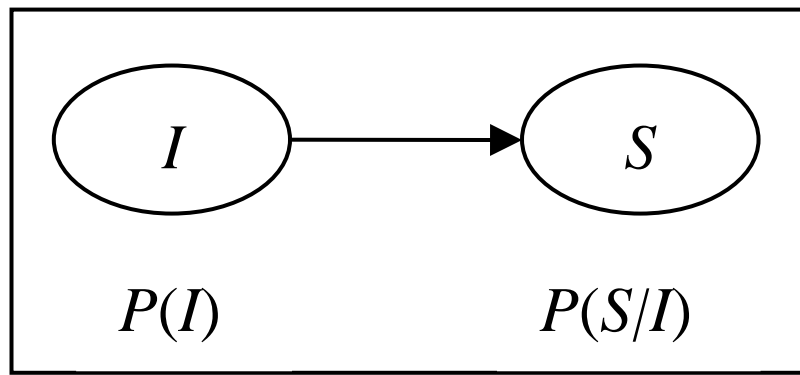

**Fig. 2**: *S* is dependent on *I*

The primary probability distribution $P(I)$ and the CPD $P(S|I)$ are given in Fig. 7 (p. 12). The joint distribution can be computed by using Bayes' chain rule:

$$P(I,S) = P(I)\,P(S|I) \tag{20}$$

It equals the product of the distributions at each node in Fig. 2. Its four numerical values can be calculated from Table 1-2. They are:

$$P(i^0,s^0) = P(i^0)\,P(s^0|i^0) = 0.7\cdot 0.95 = 0.665 \tag{21}$$

Similarly: $\{P(i^0,s^1), P(i^1,s^0), P(i^1,s^1)\} = \{0.035, 0.06, 0.24\}$

It is easy to check numerically that we have a consistent MPD of $P(I)$:

$$P(i^0) = \sum_s P(i^0, s) = 0.665 + 0.035 = 0.7, \;\; P(i^1) = 0.3, \sum_\mu P(i^\mu) = 1 \tag{22}$$

This consistency is not surprising because:

$$\sum_s P(i^\mu, s) = \sum_s P(i^\mu) P(s \mid i^\mu) = P(i^\mu) \sum_s P(s \mid i^\mu) \underset{(18)}{=} P(i^\mu) \tag{23}$$

In the same way, we can derive the MPD of $P(S)$

$$P(s^\mu) = \sum_\rho P(i^\rho, s^\mu) = \sum_\rho P(i^\rho) P(s^\mu \mid i^\rho) \tag{24}$$

Their numerical values are calculated as follows, which are also consistent.

$$P(s^0) = \sum_{\rho=1}^{2} P(i^\rho, s^0) = 0.665 + 0.06 = 0.725, \; P(s^1) = 0.275; \quad \sum_{\mu=1}^{2} P(s^\mu) = 1 \tag{25}$$

Note that Eq. (24) recovers the *P*-identity $I_I$ for $(\Omega_I, I, P)$:

$$\begin{aligned} &P(s^\mu) = \sum_\rho P(i^\rho, s^\mu) = \sum_\rho P(i^\rho) P(s^\mu \mid i^k) = \sum_\rho P(s^\mu \mid i^\rho) P(i^\rho \mid \Omega) = P(s^\mu \mid I_I \mid \Omega) \\ &\therefore \;\; P(s^\mu) = P(s^\mu \mid \Omega) = P_{S|\Omega}(s^\mu \mid \Omega) = P(s^\mu \mid I_I \mid \Omega), \quad I_I = \sum_\rho | \, i^\rho) P(i^\rho \, | \end{aligned} \tag{26}$$

By definition, the CPD $P(I \mid S)$ and its numerical values are given by:

$$P(I \mid S) = P(I, S) / P(S) \tag{27}$$

$$P(i^0 \mid s^0) = P(i^0, s^0) / P(s^0) = 0.665 / 0.725 = 0.917 \tag{28}$$

Similarly: $\{P(i^1 \mid s^0), P(i^0 \mid s^1), P(i^1 \mid s^1)\} = \{0.083, 0.127, 0.873\}$

Using Eq. (7) and (20), we can recalculate $P(I)$, matching Eq. (22):

$$P(i^0) = \sum_\mu P(i^0, s^\mu) = \sum_\mu P(s^\mu) P(i^0 \mid s^\mu) = 0.665 + 0.035 = 0.7$$

$$P(i^1) = \sum_\mu P(i^1, s^\mu) = \sum_\mu P(s^\mu) P(i^1 \mid s^\mu) = 0.06 + 0.24 = 0.3$$

We can also re-derive the *P*-identity $I_S$ in $(\Omega_S, S, P)$ as follows:

$$\begin{aligned} &P(i^\rho) = \sum_\mu P(i^\rho, s^\mu) = \sum_\mu P(i^\rho \mid s^\mu) P(s^\mu) = \sum_\mu P(i^\rho \mid s^\mu) P(s^\mu \mid \Omega) = P(i^\rho \mid I_S \mid \Omega) \\ &\therefore \;\; P(i^\rho) = P(i^\rho \mid \Omega) = P_{I|\Omega}(i^\rho \mid \Omega) = P(i^\rho \mid I_S \mid \Omega), \quad I_S = \sum_\mu | \, s^\mu) P(s^\mu \, | \end{aligned} \tag{29}$$

**2.1.3. The first Restriction on Inserting *P*-identities**: The two *P*-identities ($I_I$ and $I_S$) are derived from the expressions of cross-domain probability distributions. Therefore, they can be inserted into cross-domain probability brackets as follows:

$$P(i^k) = P(i^k \mid \Omega) = P(i^k \mid I_S \mid \Omega) = \sum_{\mu} P(i^k \mid s^\mu) P(s^\mu \mid \Omega) = \sum_{\mu} P(i^k, s^\mu) P(s^\mu)$$
$$P(s^\mu) = P(s^\mu \mid \Omega) = P(s^\mu \mid I_I \mid \Omega) = \sum_{\rho} P(s^\mu \mid i^\rho) P(i^\rho \mid \Omega) = \sum_{\rho} P(i^\rho, s^\mu)\, P(i^\rho) \tag{30}$$
$$P(i^k \mid s^\nu) = P(i^k \mid I_S \mid s^\nu) = \sum_{\mu} P(i^k \mid s^\mu) P(s^\mu \mid s^\nu) = \sum_{\mu} P(i^k \mid s^\mu) \delta_{\mu\nu} = P(i^k \mid s^\nu)$$

Of course, $I_I\,(I_S)$ can also be used to manipulate probabilities in its domain $\Omega_I\,(\Omega_S)$ of a single variable *I* (*S*). Eqs. (18, 26, 29) show that a multivariable system's JPD, IPD, and MPD are cross-domain *P*-brackets. But there are restrictions. First of all, one cannot *insert $I_Y$ into a basis P-bracket in $\Omega_X$:*

$$\delta_{ik} = P(x_i \mid x_k) \neq P(x_i \mid I_Y \mid x_k) = \sum_{\mu} P(x_i \mid y_\mu) P(y_\mu \mid x_k) \tag{31}$$

We can validate the inequality of Eq. (31) for $P(x_i, y_\mu) > 0$ as follows:

If $i \neq k,\ 0 = P(x_i \mid x_k)$, while: $P(x_i \mid I_Y \mid x_k) = \sum_{s} P(x_i \mid y_s) P(y_s \mid x_k) > 0$

If $i = k,\ 1 = P(x_i \mid x_i)$, while: $P(x_i \mid I_Y \mid x_i) = \sum_{s} P(x_i \mid y_s) P(y_s \mid x_i) < \sum_{s} P(y_s \mid x_i) \underset{(2.1.19)}{=} 1$.

In **§3.4**, we will deliver *a general rule* for inserting P-identities between nodes.

**2.1.4. Bridged expectation in PBN**: Assuming $A_X \subseteq \Omega_X, B_y \subseteq \Omega_y$, we have:

$$P(A_X \mid y) = P(A_X \mid I_X \mid y) = \sum_{x} P(A_X \mid x) P(x \mid y) = \sum_{x \in A_X} P(x \mid y) \tag{32}$$

$$P(y \mid B_X) = P(y \mid I_X \mid B_X) = \sum_{x} P(y \mid x) P(x \mid B_X) = \sum_{x \in B_X} P(y \mid x) P(x) / P(B_X) \tag{33}$$

$$P(A_X \mid Y \mid B_Y) = \sum_{y} y P(A_X \mid y) P(y \mid B_Y) \underset{(32)}{=} \sum_{x \in A_X, y \in B_X} y P(x, y) / P(B_Y) \tag{34}$$

$$P(A_x \mid B_Y) = P(A_x \mid I_Y \mid B_Y) = \sum_{y} P(A_X \mid y) P(y \mid B_Y) \underset{(32)}{=} \sum_{\substack{x \in A_X \\ y \in B_Y}} P(x, y) / P(B_Y) \tag{35}$$

Hence, the *bridged expectation* can be defined as:

$$E[A_X \mid Y \mid B_Y] \equiv P(A_x \mid Y \mid B_Y) / P(A_X \mid B_Y) = P(B_Y \mid Y \mid A_x) / P(B_Y \mid A_x) \tag{36}$$

Note that PBN can construct legitimate and useful bridged expressions, such as Eq. (34, 36), that cannot be directly identified with conventional probability notations.

## 2.2 Three-Variable Systems and Dependencies of Three-Node Chains

Suppose we have three discrete random variables, *X*, *Y*, and *Z*, each with its respective sample space. They form a joint sample space $\Omega = \{\Omega_X, \Omega_Y, \Omega_Z\}$, having the following *P*-basis for any event $\{x, y, z\} \in \{\Omega_X, \Omega_Y, \Omega_Z\}$:

$$X \mid x, y, z) = x \mid x, y, z), \quad Y \mid x, y, z) = y \mid x, y, z),$$
$$Z \mid x, y, z) = z \mid x, y, z), \quad P(\Omega \mid x, y, z) = 1 \tag{37}$$

$$\begin{aligned} &\text{Orthonormality: } P(x,y,z'\,|\,x',y',z') = \delta_{xx'}\,\delta_{yy'}\,\delta_{zz'} \\ &\text{Completeness: } \sum_{x,y,z\in\Omega} |\,x,y,z)P(x,y,z\,| \;= I_{X,Y,Z} \end{aligned} \tag{38}$$

**2.2.1. The definitions and Properties of Three-Variable PDs:** The full joint probability distribution (JPD) is defined by:

$$\begin{aligned} &P(x,y,z) = P(x,y,z\,|\,\Omega) = P(X=x, Y=y, Z=z\,|\,\Omega),\ \{x,y,z\}\in\Omega \\ &P(x,y,z)\geq 0, \text{ if } \{x,y,z\}\in\Omega; \quad P(x,y,z)=0, \text{ if } \{x,y,z\}\notin\Omega \end{aligned} \tag{39}$$

In addition to the properties and definitions we have seen for two observables, there are CPDs related to three observables. For example, here is the CPD over *X* and *Y* given *Z*:

$$P(x,y\,|\,z) \equiv P(X=x, Y=y\,|\,Z=z) = P_{X,Y|Z}(x,y\,|\,z) \tag{40}$$

**Definition 2.2.1**: Event $X = x$ is conditionally independent of event $Y = y$ given $Z= z$ and $Y = y$ in a distribution *P* (§2.1.4.2 of [3]), denoted

$$P \models (x \perp y\,|\,z), \text{ if } \; P(x\,|\,y\wedge z) \equiv P(x\,|\,y,z) = P(x\,|\,z) \tag{41}$$

**Definition 2.2.2**: Random variables *X* and *Y* are conditionally independent given variable *Z* in a distribution *P*, denoted by if $P \models (x \perp y\,|\,z)$ for $\forall x,y,z\in\Omega$. (42)

**Proposition 2.2.1**: $P \models (X \perp Y\,|\,Z)$ if and only if $P(X,Y\,|\,Z) = P(X\,|\,Z)P(Y\,|\,Z)$. (43)

Proof ($\Rightarrow$): $P(x,y\,|\,z) = P(x,y,z)/P(z) = P(x\,|\,y,z)P(y,z)/P(z)$

$= P(x\,|\,y,z)P(y\,|\,z) \underset{(41)}{=} P(x\,|\,z)P(y\,|\,z)$

($\Leftarrow$): $P(x\,|\,y,z) = P(x,y,z)/P(y,z) = P(x,y\,|\,z)P(z)/P(y,z)$

$\underset{(43)}{=} P(x\,|\,z)P(y\,|\,z)/P(y/z) = P(x\,|\,z)$ QED

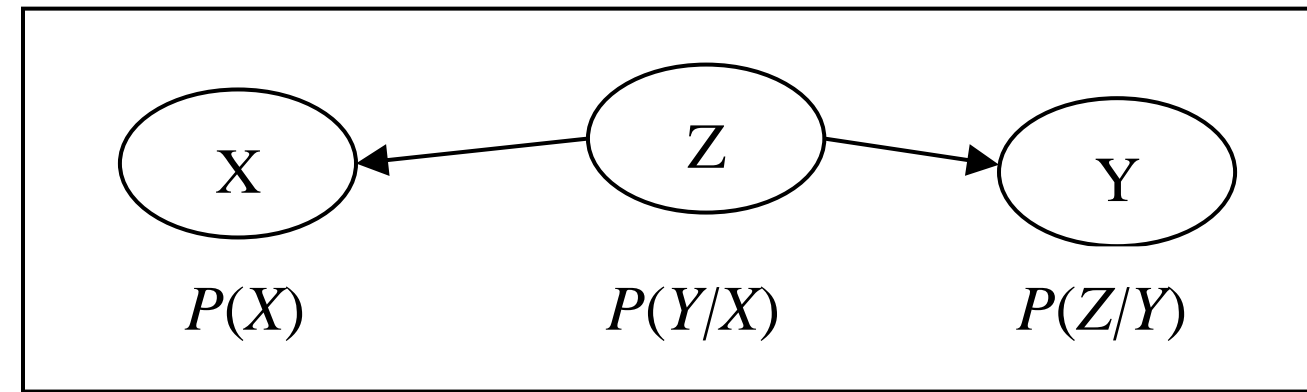

**Fig. 3**: Common Cause: $P \models (X \perp Y\,|\,Z)$

If $P \models (X \perp Y\,|\,Z)$, then using Bayes' rule and Eq. (43), we get the following JPD, as in Fig. 3.

$$P(x,y,z) = P(x,y\,|\,z)P(z) = P(x\,|\,z)P(y\,|\,z)P(z) \tag{44}$$

Fig. 4 and Bayes' chain rule produce the following two JPDs:

$$P(x,y,z) = P(x)P(y\,|\,x)P(z\,|\,y), \quad P(x,y,z) = P(x)P(y\,|\,x)P(z\,|\,x,y) \tag{45}$$

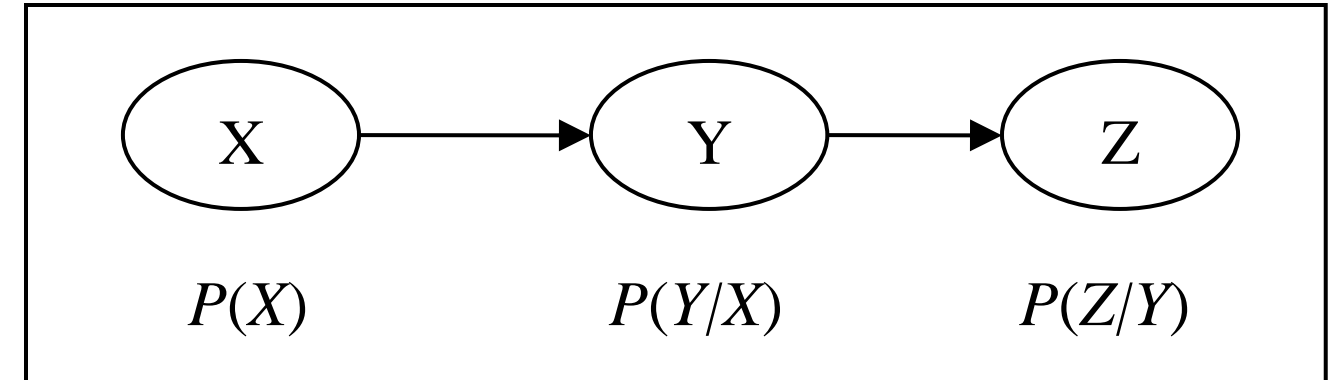

**Fig. 4**: Causal Chain: *Y* depends on *X,* and *Z* depends on *Y*

Comparing the two expressions, we get a conditional independence:

$$P(z \mid x, y) = P(z \mid y) \Rightarrow P \models (Z \perp X \mid Y) \tag{46}$$

**2.2.2. The *D*→*G*←*I* example of the Student Bayesian network** [3]: It has three variables: *D* (the difficulty of a course), *I* (the intelligence of a student), and *G* (the student's grade in the course), as shown in Fig. 5. The primary MPDs of *P*(*I*) and *P*(*D*) as well as the CPD of $P(F \mid I, D)$ are given in Fig. 7.

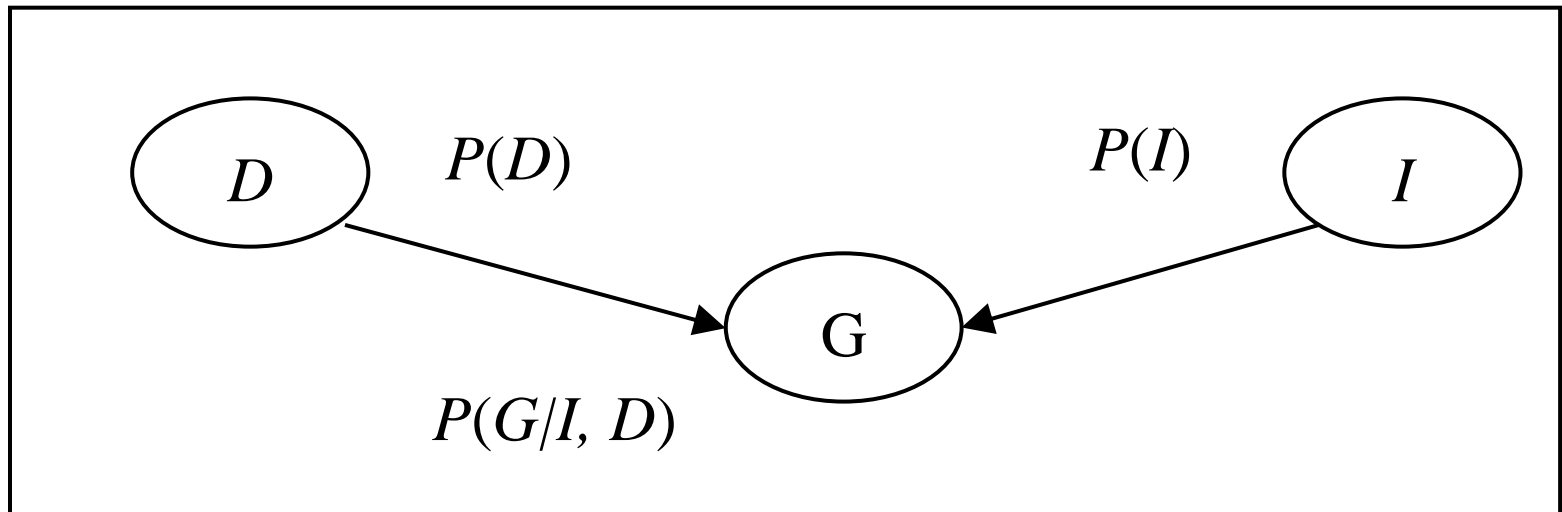

**Fig. 5**: Collider: *G* depends on *D* and *I*

Fig. 5 and the chain rule provide the following two JPDs, respectively:

$$P(d,i,g) = P(d)P(i)P(g \mid d,i), \quad P(d,i,g) = P(d)P(i \mid d)P(g \mid d,i) \tag{47}$$

Comparing the two expressions, we find:

$$P(i \mid d) = P(i) \Rightarrow P \models (D \perp I) \text{ or } P(d,i) = P(d)P(i) \tag{48}$$

From Eq. (47), we can calculate miscellaneous intermediate JPD and MPD, like:

$$\begin{aligned} P(d,g) &= \sum\nolimits_i P(d,i,g) = \sum\nolimits_i P(d)P(i)P(g \mid d,i) \\ P(d) &= \sum\nolimits_g P(d,g) = \sum\nolimits_{i,g} P(d)P(i)P(g \mid d,i) = \sum\nolimits_i P(d)P(i) = P(d) \\ P(g) &= \sum\nolimits_{i,d} P(g,i,d) = \sum\nolimits_{i,d} P(g \mid i,d)P(i)P(d) \end{aligned} \tag{49}$$

Here, we have used the CP normalization, as in Eq. (18). We can rederive *P*(*d*, *g*) in Eq. (49) using the *P*-identity $I_I$:

$$P(d,g) = P(d,g \mid \Omega) = \sum\nolimits_i P(d,g \mid i)P(i \mid \Omega) = \sum\nolimits_i \frac{P(d,i,g)}{P(i)} P(i) = \sum\nolimits_i P(d,i,g) \tag{50}$$

We can also derive *P*(*i*, *g*) by using the *P*-identity $I_D$:

$$P(i,g) = P(i,g \mid \Omega) = P(i,g \mid I_D \mid \Omega) = \sum_d P(i,g \mid d)P(d \mid \Omega) = \sum_d P(d,i,g) \quad (51)$$

Similarly, the MPD of *G* can be obtained by inserting the *P*-identity $I_{I,D}$:

$$P(g) = P(g \mid I_{I,D} \mid \Omega) = \sum_i P(g \mid i,d)P(i,d \mid \Omega) = \sum_i P(g \mid i,d)P(i)P(d) \quad (52)$$

Various CPDs can be derived from Eq. (47-53), such as:

$$P(i \mid g) = \frac{P(i,g)}{P(g)} = \frac{1}{p(g)}\sum_d P(i,d,g) = \frac{1}{p(g)}\sum_d P(g \mid i,d)P(i)P(d)$$

**Second Restriction on inserting P-identities**: Node G is called a *Collider* in the network D→G←I because D and I become dependent on G (i.e., G is activated, a case of explaining away):

$$P(d \mid i,g) = \frac{P(d,i,g)}{P(i,g)} \neq P(d \mid g) = \frac{P(d,g)}{P(g)} = \frac{\sum_i P(d,i,g)}{\sum_{i'} P(i',g)} \quad (53)$$

The conditional independence D$\perp$I|G does not hold, Eq. (41) is not valid, therefore:

$$P(i \mid d) = \sum_g P(i,g \mid d) = \sum_g P(i \mid g,d)P(g \mid d) \neq \sum_g P(i \mid g)P(g \mid d) \quad (54)$$

So far, we have seen the three important dependencies in 3-node networks: *Common cause* (Fig. 3), *Causal chain* (Fig. 4), and *Collider* (Fig. 5). We will revisit these dependencies in **§3.4** when we present a *general rule* for inserting P-identities between nodes.

**Conditional expectations** (CEs) are widely used in statistical analysis. Here is an example of dual-conditional expectations for the $D \rightarrow G \leftarrow I$ subnet:

$$\begin{aligned} E[G \mid d,i] &\equiv P(\Omega \mid G \mid d,i) = P(\Omega \mid GI_G \mid d,i) = \sum_g P(\Omega \mid G \mid g)P(g \mid d,i) \\ &= \sum_g gP(g \mid d,i) = \sum_g gP(g,d,i)/P(d,i) \end{aligned} \quad (55)$$

## 2.3 Definitions and Rules on Bayesian Networks (Informal)

Here, we introduce some informal graphic definitions and rules for BN**.** Please refer to Refs. [3], [4], and [5] for more details. The following informal definitions and regulations are quoted from Lecture 2 of Ref. [5].

**The Bayesian network (informal)**

- A directed acyclic graph (DAG) G
- Nodes represent random variables
- Edges represent direct influences between random variables
- Local probability models (conditional parameterization)
- Conditional probability distributions (CPDs)

**The Bayesian network structure**

- A directed acyclic graph (DAG) G

- Nodes $X_1, \ldots, X_n$ represent random variables
- G encodes the following set of independence assumptions (called local independencies)
    - $X_i$ is independent of its non-descendants, given its parents
    - Formally: $(X_i \perp \text{NonDesc}(X_i) \mid P_a(X_i))$ (56)
    - Denoted by $I_L(G)$

**3. The Student Bayesian Network and Reasoning Examples**

We are ready to examine the renowned Student Bayesian Network [3-5]. In §3.1, we will present the Probabilistic Graphic Model (PGM, [6]), local dependencies, the probability distribution (PD) for each node, and some reasoning results. Following that, in §3.2, we will complete our reasoning problems to derive these reasoning results. Then, in §3.3, we will discuss how to use PBN to unify and simplify calculations related to various probability distributions for predictive and reasoning tasks.

**3.1. Local Independences of Student BN and Some Reasoning Results**

The Student BN [3-5] has five random variables with their sample spaces, $\{\Omega_D, \Omega_I, \Omega_G, \Omega_S, \Omega_L\}$ as shown in Fig. 6. Applying the rules in §2.3, specifically, Eq. (56), we obtain the following local independences [5] for Student BN as $I_L(G_{student})$:

$$
\begin{aligned}
&D \perp I; \quad D \perp S; \quad G \perp S \mid D, I \\
&L \perp I, D, S \mid G; \qquad S \perp D, G, L \mid I
\end{aligned}
\tag{57}
$$

To be consistent with Ref. [3], we list the outcomes and associated PD of the five nodes of the Student BN in Fig. 6 with numerical values[2] as follows:

$$
\begin{aligned}
&\text{Intelligence:} \quad \text{Val}(I) = \{i^0, i^1\} = \{\text{low}, \text{high}\} = \{0,1\} = \Omega_I \\
&\text{Difficulty:} \quad \text{Val}(D) = \{d^0, d^1\} = \{\text{easy}, \text{hard}\} = \{0,1\} = \Omega_D \\
&\text{Grade:} \quad \text{Val}(G) = \{g^1, g^2, g^3\} = \{A, B, C\} = \{4,3,2\} = \Omega_G \\
&\text{SAT:} \quad \text{Val}(S) = \{s^0, s^1\} = \{\text{low}, \text{high}\} = \{0,1\} = \Omega_S \\
&\text{Letter:} \quad \text{Val}(L) = \{l^0, l^1\} = \{\text{weak}, \text{strong}\} = \{0,1\} = \Omega_L
\end{aligned}
\tag{58}
$$

The PGM in Fig. 6 tells us about the PD associated with each node:

$$
\{D, I, G, S, L\} \Rightarrow \{P(d), P(i), P(g \mid d, i), P(s \mid i), P(l \mid g)\} \tag{59}
$$

It also tells us the expression of the full joint probability distribution:

$$
P(d, i, g, s, l) = P(d)P(i)P(g \mid d, i)P(s \mid i)P(l \mid g) \tag{60}
$$

---

[2] Here, we set numerical values of the variables for calculating their expectation values in Section **3.4**

On the other hand, using the chain rule, we have

$$P(d,i,g,s,l) = P(d)P(i \mid d)P(g \mid d,i)P(s \mid d,i,g)P(l \mid d,i,g,s) \tag{61}$$

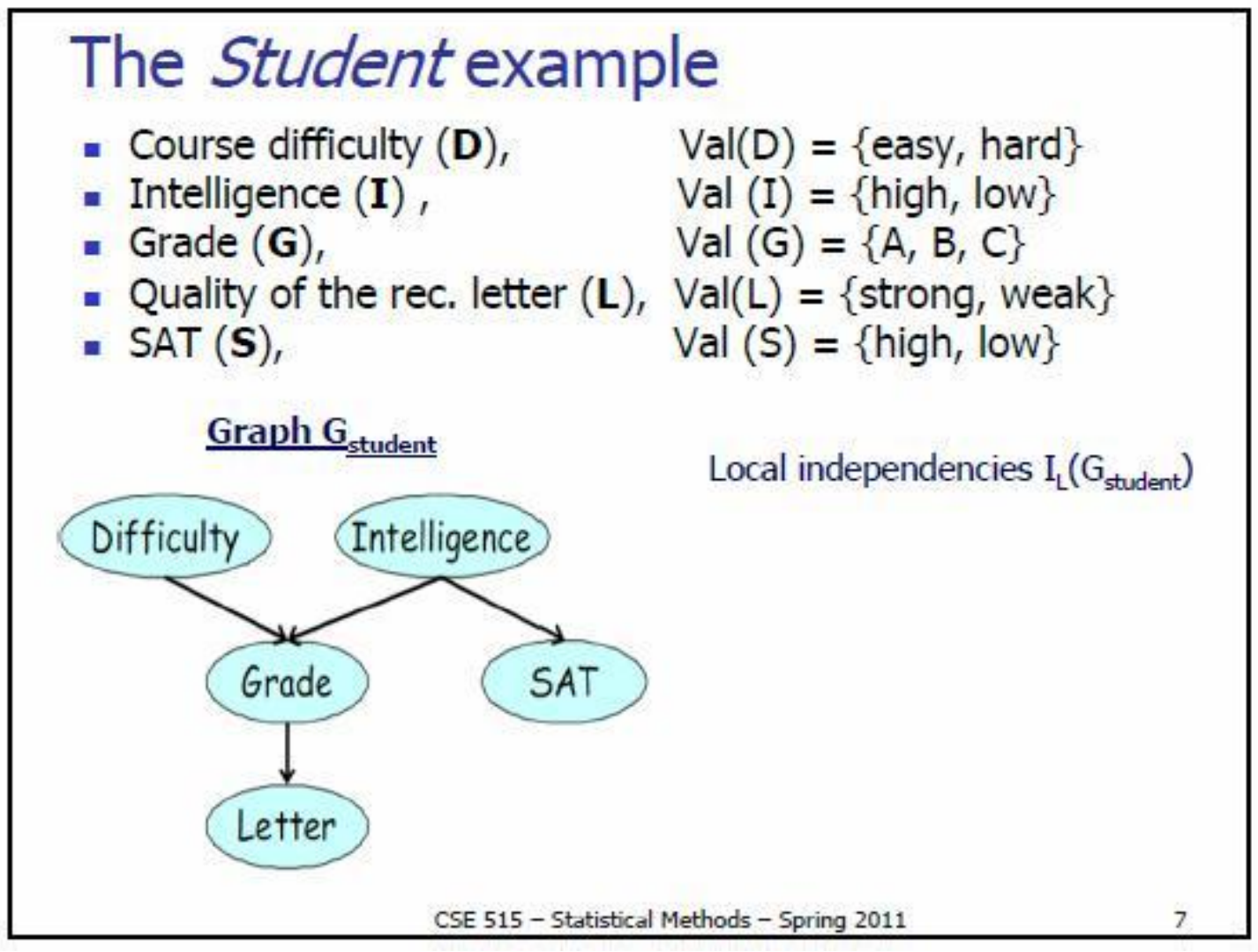

**Fig. 6**: The Student Bayesian Network Example (see Ref [5])

We can use relations in Eq. (57) to derive Eq. (60) from Eq. (61), but as one can see, it is much easier to get Eq. (60) by using PGM for the BN.

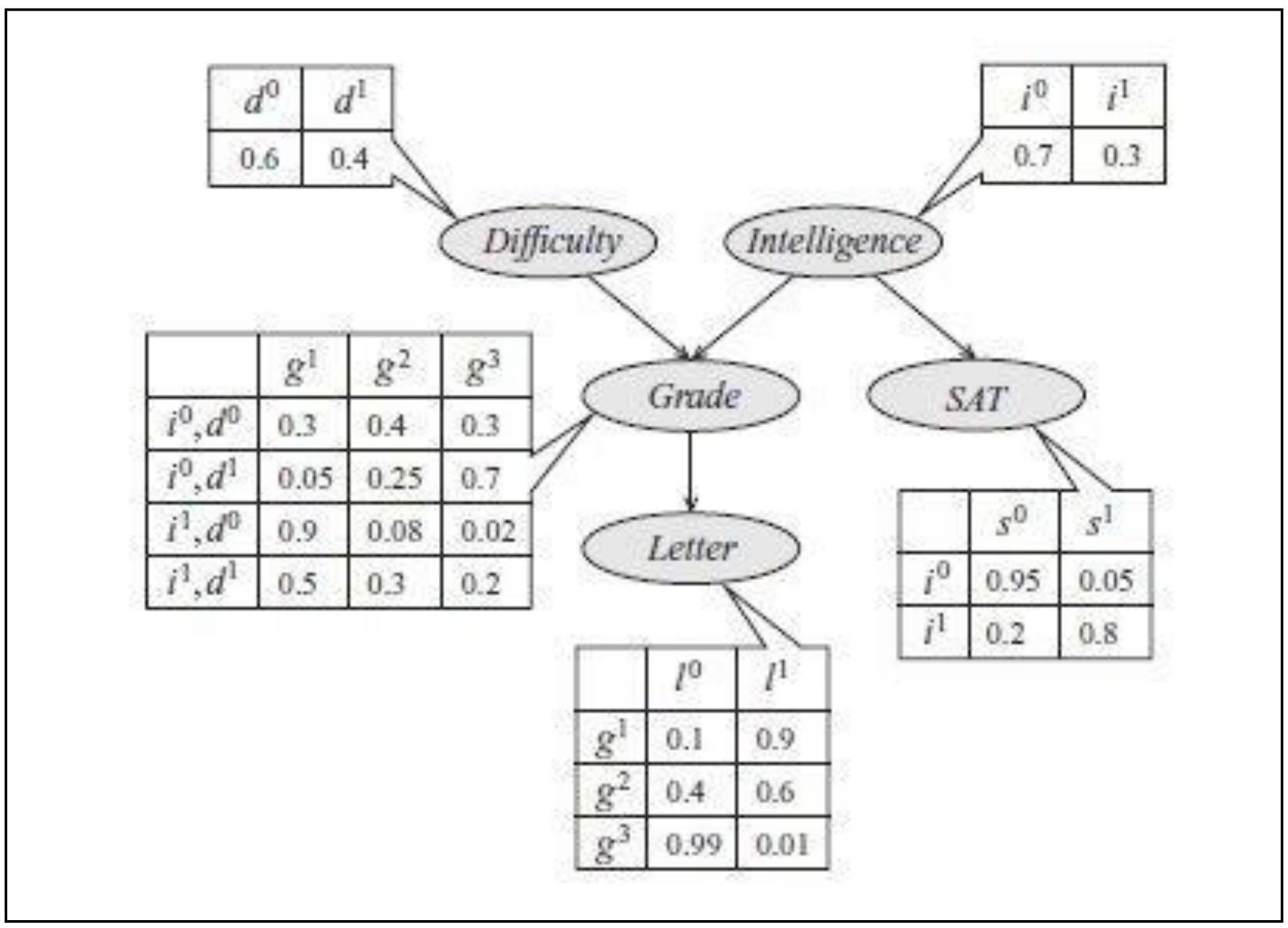

| $d^0$ | $d^1$ |
| --- | --- |
| 0.6 | 0.4 |

| $i^0$ | $i^1$ |
| --- | --- |
| 0.7 | 0.3 |

| | $g^1$ | $g^2$ | $g^3$ |
| --- | --- | --- | --- |
| $i^0,d^0$ | 0.3 | 0.4 | 0.3 |
| $i^0,d^1$ | 0.05 | 0.25 | 0.7 |
| $i^1,d^0$ | 0.9 | 0.08 | 0.02 |
| $i^1,d^1$ | 0.5 | 0.3 | 0.2 |

| | $s^0$ | $s^1$ |
| --- | --- | --- |
| $i^0$ | 0.95 | 0.05 |
| $i^1$ | 0.2 | 0.8 |

| | $l^0$ | $l^1$ |
| --- | --- | --- |
| $g^1$ | 0.1 | 0.9 |
| $g^2$ | 0.4 | 0.6 |
| $g^3$ | 0.99 | 0.01 |

**Fig. 7**: The PD for the Student BN ([3], [4])

The numerical values of the PD tables for the five nodes in Eq. (60) are presented in Fig. 7 ([3], [4]). To simplify our statements, we assume that the student is named George and the course name is Econ101 ([3], Sec. 3.2). From the Student BN, we can then make direct predictions such as:

1. If George has a grade $g^1 = A$, then the probability of him getting a strong letter from his professor is $P(l^1 \mid g^1) = 0.9$, independent of the values of *D*, *I,* or *S*.
2. If George has high intelligence ($i^1$) and the course is hard ($d^1$), then the probability of him getting a grade $g^1 = A$ is $P(g^1 \mid i^1, d^1) = 0.5$, independent of *S*.
3. If George has low intelligence ($i^0$), then the probability of him getting a high SAT ($s^1$) is $P(s^1 \mid i^0) = 0.05$, independent of *D*, *G,* or *L*.

Another critical aspect of the BN is its reasoning capabilities (top-down or bottom-up reasoning, prediction, explanation, or inference [3]). Here are some reasoning problems (**RP**s) of the Student BN, many of which are provided in Ref. [3] without derivation. In the next section, we will solve them using the PD tables shown in Fig. 7.

**RP-1**: Assuming we know nothing about Econ101 and George. Figure 3.1.2 shows that George has a 30% chance of being highly intelligent $P(i^1) = 0.3$. However, if we know his Econ 101 grade is $g^3$ = C, his chance of being high-intelligence is reduced to 7.9% (as an example of inference in bottom-up reasoning).

**RP-2**: Assuming we know nothing about Econ101 and George, he has a 36.2% chance of getting a grade A, i.e., $P(g^1) = 0.362$. But if we know that George has low intelligence, the probability that he will get a grade A is reduced to 20% (as in a top-down prediction).

**RP-3**: Assuming we know nothing about Econ101 and George, he will have a 50.2% chance of getting a strong recommendation letter, i.e., $P(l^1) = 0.502$. But if George has low intelligence $(i^0)$, his chance of getting a strong letter will be reduced to 38.9%, or $P(l^1 \mid i^0) = 0.389$. Now, if we also know that the course is easy, his chance of getting a strong letter increases to 51.3% (0.513), as a case of explaining away in top-down reasoning.

### 3.2 Deriving the Reasoning Results for the Student BN

To complete our reasoning, we must use various marginal and conditional PDs, defined in or derived from the tables in Fig. 7.

#### 3.2.1: The Marginal PD of *G* (Grade) or $P(g) = P(g \mid \Omega)$:

Because we know $P(g \mid i, d)$ and $P(i, d \mid \Omega) = P(i, d) = P(i)P(d)$, it is natural to insert the identity operator $I_{I,D}$, as we did in Eq. (55):

$$P(g) = P(g \mid \Omega) = P(g \mid I_{I,D} \mid \Omega) = \sum_{i,d} P(g \mid i,d) P(i,d) \\ = \sum_{i,d} P(g \mid i,d) P(i) P(d) \tag{62}$$

The traditional way to derive Eq. (62) is from the joint PD and the chain rule:

$$P(g) = \sum_{i,d} P(g,i,d) = \sum_{i,d} P(g \mid i,d) P(i,d)$$

It is now straightforward to evaluate $P(g)$:

$$P(g^{\mu}) = \sum_{\alpha,\beta} P(g^{\mu}, i^{\alpha}, d^{\beta}) = \sum_{\alpha,\beta} P(g^{\mu} \mid i^{\alpha}, d^{\beta}) P(i^{\alpha}) P(d^{\beta}) \quad (63)$$

Using tables in Fig. 7 and Eq. (63), we have the following MPD with valid normalization.

$P(g^1) = 0.3620$ (As inRP-2), $P(g^2) = 0.2884$, $P(g^3) = 0.3496$; $\sum_g P(g) = 1$

**3.2.2: The Marginal PD of *L* (Letter) or** $P(l) = P(l \mid \Omega)$: Because we already know $P(l \mid g)$ and $P(g \mid \Omega) = P(g)$, it is natural to insert the identity operator $I_G$:

$$P(l) = P(l \mid \Omega) = P(l \mid I_G \mid \Omega) = \sum_g P(l \mid g) P(g \mid \Omega) = \sum_g P(l \mid g) P(g) \quad (64)$$

Equivalently, it can be derived by using Bayes' chain rule as:

$$P(l) = \sum_g P(l, g) = \sum_g P(l \mid g) P(g) \quad (65)$$

Using tables in Fig. 7 and Eq. (65), it is straightforward to find: $P(l^1) \approx 0.502$, as in RP-3.

**3.2.3: The Conditional PD of *I* (Intelligence) given *G* or** $P(i \mid g)$: The expression has already been derived in Eq. (54):

$$P(i \mid g) = P(i, g) / P(g) = \sum_d P(g \mid i, d) P(i) P(d) / P(g)$$

Using tables in Fig. 7 and Eq. (52), it is easy to obtain:

$P(i^1 \mid g^3) \approx 0.079$ (As in RP-1)

**3.2.4: The Conditional PD of *L* (Letter) given *I* and *D* or** $P(l \mid i, d)$:

$$P(l \mid i, d) = P(l \mid I_G \mid i, d) = \sum_g P(l \mid g) P(g \mid i, d) \quad (66)$$

Using tables in Fig. 7, one gets: $P(l^1 \mid i^0, d^0) \approx 0.513$ (As in RP-3)

**3.2.5: The Conditional PD of *G* (Grade) given *I* or** $P(g \mid i)$:

$$P(g \mid i) = \frac{P(g, i)}{P(i)} = \frac{\sum_d P(g, i, d)}{P(i)} = \frac{\sum_d P(g \mid i, d) P(i, d)}{P(i)} = \sum_d P(g \mid i, d) P(d) \quad (67)$$

Assuming that George has low intelligence and using the tables in Fig. 7, we have:

$P(g^1 | i^0) = 0.20$ (As inRP-2), $P(g^2 | i^0) = 0.34$, $P(g^3 | i^0) = 0.46$

Again, the CP normalization is satisfied:

$$1 = P(\Omega | i^0) = P(\Omega | I_G | i^0) = \sum_g P(g | i^0) = 1.00$$

**3.2.6: The Conditional PD of *L* (Letter) given *I* or** $P(l | i)$ :

$$P(l | i) = P(l | I_G | i) = \sum_g P(l | g) P(g | i) \tag{68}$$

Using the table $P(l | g)$ in Fig. 7 and Eq. (68), we obtain:

$P(l^1 | i^0) = \sum_g P(l^1 | g) P(g | i^0) \approx 0.389$ (As in RP-3)

So far, we have completed the reasoning exercises for Student BN, as outlined in §3.1. Our findings show that various probabilistic expressions can be derived by simply inserting the suitable *P*-identity operators. Various software packages exist for building and testing BNs [8]. To keep the scope of this article, we will not discuss or use them.

### 3.3. Manipulating Probabilistic Expressions in Static BNs with PBN

In the previous subsections, we introduced the Student BN and showed how the Probability Bracket Notation (PBN) framework can be used to solve inference problems in a structured way. While traditional Bayesian network toolboxes, like the Bayes Server [14], are designed to automate these calculations, PBN offers a more flexible operator-based language with a clear mechanism for simplifying the analytical calculations of prediction, reasoning, and expectation in multivariable systems. The main advantage of PBN is its use of operators (observables and *P*-identities) to compute such predictive quantities given the full joint probability distribution function (JPDF) decomposed by the directed acyclic graph (DAG), as Eq. (60) for the Student BN:

$$P(d, i, g, s, l) = P(d) P(i) P(g | d, i) P(s | i) P(l | g) \tag{69}$$

Note that, from the DAG or the decomposition, nodes *D* and *P* are independent:

$$P(d, i) = P(d) P(i) \tag{70}$$

**3.3.1. Predicted Marginal PDs of Variables and Their Expectation Values**: From Eq. (69), the MPDs for *D* and *I* are primary. Their numerical values are given by Eq. (58), and their unconditional expectation values can be readily calculated.

$$E[D] \equiv \langle D \rangle = P(\Omega | D | \Omega) = P(\Omega | D\, I_D | \Omega) = \sum_d d\, P(d) = 0.4$$
$$E[I] \equiv \langle I \rangle = P(\Omega | I | \Omega) = P(\Omega | I\, I_I | \Omega) = \sum_i i\, P(i) = 0.3$$

For G, we already calculated its MPD in Eq. (62) as follows:

$$P(g) = P(g\,|\,\Omega) = P(g\,|\,I_{I,D}\,|\,\Omega) = \sum_{i,d} P(g\,|\,i,d)P(i,d) = \sum_{i,d} P(g\,|\,i,d)P(i)P(d) \tag{71}$$

This leads to the (unconditional) expectation value of Grade:

$$\begin{aligned}\langle G\rangle &= P(\Omega\,|\,G\,|\,\Omega) = P(\Omega\,|\,G\,I_G\,|\,\Omega) = \sum\nolimits_g P(\Omega\,|\,G\,|\,g)P(g\,|\,\Omega) = \sum\nolimits_g gP(g\,|\,\Omega) \\ &= \sum\nolimits_g gP(g) = \sum\nolimits_{g,i,d} gP(g\,|\,i,d)P(i)P(d)\end{aligned} \tag{72}$$

On the other hand, one can also calculate it by using the full joint PD in Eq. (69):

$$\begin{aligned}\langle G\rangle &= \sum\nolimits_{d,i,g,s,l} gP(d,i,g,s,l) = \sum\nolimits_{d,i,g,s,l} gP(d)P(i)P(g\,|\,d,i)P(s\,|\,i)P(l\,|\,g) \\ &= \sum\nolimits_{g,d,i} gP(d)P(i)P(g\,|\,d,i) \underset{(71)}{=} \sum\nolimits_g gP(g)\end{aligned} \tag{73}$$

Here, we have used the normalization condition of conditional probabilities:

$$\begin{aligned}1 &= P(\Omega\,|\,i) = P(\Omega\,|\,I_S\,|\,i) = \sum\nolimits_s P(\Omega\,|\,s)P(s\,|\,i) = \sum\nolimits_s P(s\,|\,i) \\ 1 &= P(\Omega\,|\,g) = P(\Omega\,|\,I_L\,|\,g) = \sum\nolimits_l P(\Omega\,|\,l)P(l\,|\,g) = \sum\nolimits_l P(l\,|\,g)\end{aligned} \tag{74}$$

Similarly, we can calculate the marginal PDs and means of SAT and Letter as follows:

$$P(s) = P(s\,|\,I_I\,|\,\Omega) = \sum_i P(s\,|\,i)P(i), \quad \langle S\rangle = P(\Omega\,|\,S\,|\,\Omega) = \sum_{s.i} sP(s\,|\,i)P(i) \tag{75}$$

$$P(l) = P(l\,|\,I_G\,|\,\Omega) = \sum_g P(l\,|\,g)P(g\,|\,\Omega) = \sum_g P(l\,|\,g)P(g), \quad \langle L\rangle = \sum_{l.g} l\,P(l\,|\,g)P(g) \tag{76}$$

Here, we need to use $G$'s previously derived MPD in Eq. (71). Thus, the MPDs allow one to track how uncertainty propagates across the network and find expectations.

**3.3.2. Predicted Conditional PDs, Inferences, and Conditional Expectations:** PBN also provides a consistent framework to calculate conditional dependencies by splitting brackets and inserting appropriate *P*-identities. This operator-driven manipulation is particularly useful for inference problems, as it enables predicting conditional independence via intermediate nodes. All three reasoning problems are of this type. Here we redo them with PBN to demonstrate how to use definitions of PD and insert appropriate *P*-identities. Three new reasoning problems (**RP**s) have been added and analyzed.

**RP-1**: Find $P(i^1\,|\,g^3)$. It is a reverse inference (bottom-up). In addition to what is done in Eq. (54), it can also be derived using a *P*-identity:

$$\begin{aligned}P(i\,|\,g) &= P(g\,|\,I_{I,D}\,|\,i)P(i)\,/\,p(g) = \sum_{d,i'} P(g\,|\,i',d)P(i',d\,|\,i)P(i)\,/\,P(g) \\ &= \sum_{d,i'} P(g\,|\,i',d)P(i'\wedge d\wedge i)\,/\,P(g) = \sum_{d} P(g\,|\,i,d)P(i)P(d)\,/\,P(g)\end{aligned} \tag{77}$$

**RP-2**: It is a top-down reasoning. From the MPD of *G*, we know that $P(g^1) = 0.3620$. Now what is $P(g^1|i^0)$? It can be derived by inserting an identity operator and the definition of CPD, as well as the independence of *I* and *D*, in addition to Eq. (67):

$$\begin{aligned} P(g\,|\,i) &= P(g\,|\,I_{I,D}\,|\,i) = \sum_{i',d} P(g\,|\,i',d)P(i',d\,|\,i) = \sum_{i',d} P(g\,|\,i',d)P(i'\wedge d\wedge i)/P(i) \\ &= \sum_{i',d} P(g\,|\,i',d)\delta_{i,i'}P(d,i)/P(i) = \sum_{d} P(g\,|\,i,d)P(d) \end{aligned} \tag{78}$$

**RP-3**: What is $P(l|i)$? It is a case of explaining away in top-down reasoning. From the MPD of *L,* we know $P(l^1) \approx 0.502$. But if we know George's intelligence, then we need:

$$P(l\,|\,i) = \sum_{g} P(l\,|\,g)P(g\,|\,i) \underset{(78)}{=} \sum_{g} P(l\,|\,g)\sum_{d} P(g\,|\,i,d)P(d) \tag{79}$$

Now, if we also know that the course is easy, his chance of getting a strong letter will increase according to:

$$P(l\,|\,i,d) = P(l\,|\,I_G\,|\,i,d) = \sum_{g} P(l\,|\,g)P(g\,|\,i,d) \tag{80}$$

**RP-4** (new): Another top-down inference. What is George's probability of getting a grade *B* if the course difficulty is easy, i.e., $P(g^2|d^1)$? Similar to Eq. (78), we have:

$$\begin{aligned} P(g\,|\,d) &= \sum_{i,d'} P(g\,|\,i,d')P(i,d'\,|\,d) = \sum_{i,d'} P(g\,|\,i,d')P(i\wedge d\wedge d')/P(d) \\ &= \sum_{i,d'} P(g\,|\,i,d')\delta_{d,d'}P(d,i)/P(d) = \sum_{i} P(g\,|\,i,d)P(i) \end{aligned} \tag{81}$$

**RP-5** (new): Another bottom-up inference. What is the probability of an easy course if we know George's Letter is strong? We need the following CPD:

$$P(d\,|\,l) = P(l\,|\,d)\frac{P(d)}{P(l)} = \frac{P(d)}{P(l)}P(l\,|\,I_G\,|\,d) = \frac{P(d)}{P(l)}\sum_{g} P(l\,|\,g)P(g\,|\,d) \tag{82}$$

**RP-6** (new): *Conditional Expectations* (CEs). One has already been done in Eq. (55):

$$E[G\,|\,d,i] = P(\Omega\,|\,GI_G\,|\,d,i) = \sum\nolimits_{g} gP(g\,|\,d,i) = \sum\nolimits_{g} gP(g,d,i)/P(d,i)$$

We can easily obtain other CE expressions and their relations for the student SN:

$$E[G\,|\,d] \equiv P(\Omega\,|\,G\,|\,d) = P(\Omega\,|\,GI_G\,|\,d) = \sum_{g} gP(g\,|\,d) \underset{(81)}{=} \sum_{g,i} gP(g\,|\,i,d)P(i) \tag{83}$$

$$E[G\,|\,i] = P(\Omega\,|\,GI_G\,|\,i) = \sum_{g} gP(g\,|\,i) \underset{(78)}{=} \sum_{g,d} gP(g\,|\,i,d)P(d) = \sum_{d} E[G\,|\,i,d]P(d) \tag{84}$$

$$E[G] \equiv \langle G\rangle = \sum\nolimits_{g} gP(g) \underset{(71)}{=} \sum\nolimits_{g,i,d} gP(g\,|\,i,d)P(i)P(d) = \sum\nolimits_{i,d} E[G\,|\,i,d]P(i)P(d) \tag{85}$$

**3.4. General P-Identity Insertion Rule and Efficiency of Chain Inference**

We begin with the **d-separation** in probability modeling. Let X, Y, Z be three nodes in a BN, and Z is the intermediate node between X and Y. If the activated Z *d-separates* X from Y, then X⊥Y|Z, and we can insert the P-identity of Z into P($x$|$y$):

$$P(x \mid y) = P(x \mid I_z \mid y) = \sum\nolimits_z P(x \mid z) P(z \mid y) \tag{86}$$

Note: if Z is a collider, *X→Z←Y,* then X⊥Y|Z does not hold if Z is activated, as explained in Eq. (54). Hence, collider Z does not d-separate X from Y when Z is activated.

**3.4.1. General P-Identity Insertion Rule and D-Separable Chains:** If activated Z1 d-separates X from Z2, and activated Z2 d-separates Z1 from Y, then {Z1−Z2} d-separates X from Y. Recursively, we have:

**General Rule for Inserting P-Identities:** Let nodes X and Y be two conditionally dependent nodes. If the sequence of intermediate nodes {Z1− Z2−… Zk} (activated, ordered along the edge-connected path from X to Y) d-separates X from Y in the Bayesian network, then:

$$P(x \mid y) = P(x \mid I_{z1} I_{z2} ... I_{zk} \mid y) \tag{87}$$

Otherwise, inserting identities between nodes may not be valid.

**D-Separable Chains:** We call such a sequence of nodes a *d-separable chain*. We can insert P-identities to calculate a CPT, e.g., P(a|b), for chain inference if the chain is a d-separable chain from A to B. The d-separability of a chain completely depends on the structure of the directed acyclic graph (DAG). Here are some examples related to the chain's d-separability:

- **E-1**: In a 2-node network, node Y is not a d-separator of another node X: relation X⊥X|Y cannot be true, therefore, $P(x_i|x_k) \neq P(x_i|I_y|x_k)$, as shown in Eq. (31).
- **E-2**: *V-structures* (Colliders): In *D→G←I* subnet of the Student BN, collider G does not d-separate *D* from *I* when activated, inserting $I_G$ between *P*(*D*|*I*) is invalid, as shown in Eq. (54). Similarly, in a 5-node network, the chain *A→D→G←I ←B* is not d-separable.
- **E-3**: *Lambda-structures* (Common Cause): In *G←I →S* subnet of the Student BN, we can insert $I_I$ into *P*(*G*|*S*), because G⊥S|I. Likewise, in a 5- node network, the chain *A←G←I →S →B* (or its reverse) is d-separable.
- **E-4**: *Causal* (*Serial*) *chains*: one-way d-separable chains like *I→G→L* in the Student BN or *A→B→C→D →E* (or its reverse) in a 5-node network.
- **E-5**: *Loopy networks* (e.g., a network is the 6 corners of a hexagon): there may not be a single d-separable chain because multiple active paths can connect two given nodes.
- **E-6**: *Pendant Subnets*. A pendant subnet at node C is a set of nodes that has only one connection to the rest of the network via C. For a d-separable chain like A→C→B, a pendant subnet at C does not affect the d-separation: A ⊥ B | C.

**3.4.2. Chain Inference and Efficiency:** PBN can efficiently reduce the number of operations for d-separable chain inferences, compared to directly using the full joint distribution table (FJPT). In Appendix A, we show that, for a large-scale $N$-node binary *polytree* (a singly connected BN; see §2.2 in [3]), the number of operations for a d-separable chain with $k$ intermediate nodes is about $k\cdot 2^{k+2}$ after a one-time preparation. Comparing with the order of $O(N\cdot 2^N)$ using the FJDT (§A.2), the ratio is:

$$(k/N)\cdot(2^{k+2}/2^N) \sim (k/N)\, 2^{k-N} \tag{88}$$

Therefore, for d-separable chains, inferences can be efficiently computed with local operators, which, when combined with preprocessing (§A.1), significantly reduce the number of operations. The approach is particularly efficient for d-separable chains on a polytree or its extension, as defined below.

**Definition 3.4.1**. *An* ***Extended Polytree*** is a Bayesian network whose global structure is a polytree, in which nodes may be attached to *pendant blobs* (multiply connected subnets connected to the polytree backbone via a single bridge node).

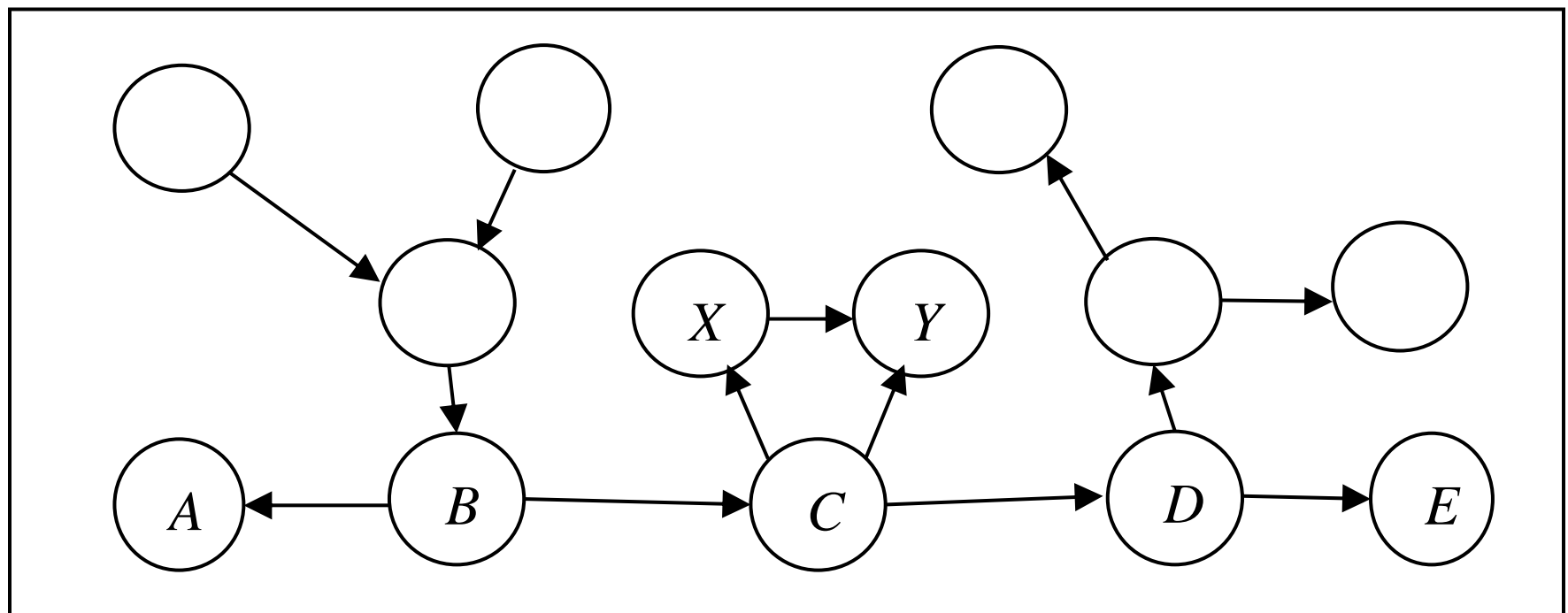

Fig. 8: An *Extended Polytree* with one pendant blob (a pendant loop):
The 3 pendant subnets do not affect the 5-node d-separable chain from A to E.

On extended polytrees, pendant subnets (including blobs) do not affect the d-separable chains on the polytree backbone. For the 5-node d-separable chain in Fig. 8, we can readily write:

$$P(a\,|\,e) = P(a\,|\,I_B I_C I_D\,|\,e) = \sum_{b,c,d} P(a\,|\,b)P(b\,|\,c)P(c\,|\,d)P(d\,|\,e)$$

Altogether, PBN serves as both a conceptual and a computational tool for d-separable chains in static Bayesian networks: it provides a structured algebraic framework for Bayesian inference using P-identities (details on the two-phase procedure are in Appendix A).

## 4. Static Bayesian Networks with Continuous and Discrete Random Variables

So far, we have discussed only static BNs for discrete random variables. However, static Bayesian Network models with continuous or mixed random variables are used in many scenarios. Examples include predicting how temperature, humidity, and pressure influence the chances of rain; a medical diagnosis system where a patient's blood pressure, heart rate, and temperature indicate the likelihood of a specific disease; and an engineering system analyzing sensor readings such as vibration levels and current to forecast possible equipment failure [9].

PBN can handle such problems easily because its probability expressions are basis-independent (either discrete or continuous) [1].

### 4.1. Continuous random variables

The expressions in Section 2 can be easily extended to continuous random variables. For example, Eq. (38) becomes:

$$\begin{aligned} &P(x,y,z\,|\,x',y',z') = \delta(x-x')\delta(y-y')\delta(z-z') \\ &\int_{\Omega} dx\,dydz\,|\,x,y,z)P(x,y,z\,| = I_{X,Y,Z} \end{aligned} \tag{89}$$

The CP normalization, similar to Eq. (18), now reads:

$$1 = P(\Omega\,|\,x,z) = P(\Omega\,|\,I_Y\,|\,x,z) = \int dy\,P(\Omega\,|\,y)P(y\,|\,x,z) = \int dy\,P(y\,|\,x,z) \tag{90}$$

Using PBN, we can readily derive many useful expectation formulas, like the biconditional expectation in Eq. (36), or the unconditional one ([2], p. 106):

$$\begin{aligned} &E[X] \equiv P(\Omega\,|\,X\,|\,\Omega) = P(\Omega\,|\,XI_X\,|\,\Omega) = \int dx\,xP(\Omega\,|\,x)P(x\,|\,\Omega) = \int dx\,xP(x\,|\,I_Y\,|\,\Omega) \\ &= \int dy \int dxxP(x\,|\,y)P(y) = \int dy\,P(\Omega\,|\,X\,|\,y)P(y) \equiv \int dyE[X\,|\,y]P(y) \equiv E[E[X\,|\,Y] \end{aligned} \tag{91}$$

The following conditional expectations will be used in the next subsection.

$$\begin{aligned} &E[Y\,|\,x,z] = P(\Omega\,|\,Y\,|\,x,z) = P(\Omega\,|\,Y\,I_Y\,|\,x,z) = \int_{y\in\Omega_y} dyP(\Omega\,|\,Y\,|\,y)P(y\,|\,x,z) \\ &= \int_{y\in\Omega_y} dyyP(y\,|\,x,z) = \int_{y\in\Omega_y} dyyP(x,y,z)\,/\,P(x,z) \end{aligned} \tag{92}$$

$$P(H_Y\,|\,x,z) = P(H_Y\,|\,I_Y\,|\,x,z) = \int dyP(H_Y\,|\,y)P(y\,|\,x,z) = \int_{y\in H_Y} dyP(x,y,z)\,/\,P(x,z) \tag{93}$$

The d-separable chains and the general identity-inserting rule discussed in **§3.4** can also be applied analytically to nodes of continuous random variables. One only needs to use the continuous representation of their identity operators in expressions like Eq. (87) and to replace summations with integrals in expressions like Eq. (86).

### 4.2. Linear Gaussian Networks and Discrete-Display Nodes

In a Bayesian network, the continuous random variables are usually assumed to follow a Gaussian (normal) distribution (with their mean and variance estimated from the given dataset).

$$P(x) = (2\pi\sigma^2)^{-1/2} \exp\{-(x-\mu)^2\,/\,2\sigma^2\} \equiv N(x\,|\,\mu,\sigma^2) \tag{94}$$

It is also Gaussian if a variable has multiple parents, each with a normal distribution. Assume $\boldsymbol{X} = \{X_1,.., X_k\}$ are jointly Gaussian with the following distribution [10-12]:

$$P(\boldsymbol{x}) = N(\boldsymbol{x} \mid \boldsymbol{\mu}, \Sigma) = (2\pi)^{-k/2} \mid \Sigma \mid^{-1/2} \exp\{-\sum_{i,j=1}^{k} (x_i - \mu_i)^T \Sigma^{-1}{}_{i,j} (x_j - \mu_j) \qquad (95)$$

$$\Sigma_{i,j} \equiv E[(x_i - \mu_i)(x_j - \mu_j)] = E[x_i x_j] + \mu_i \mu_j$$

If we set $X_k = Y$ and $\boldsymbol{X} = \{X_1,.., X_{k-1}\}$, we obtain the conditional distribution of $P(Y|\boldsymbol{X})$ by definition from the joint distribution and marginal distribution:

$$P(y \mid x_1, \ldots, x_{k-1}) \equiv P(Y \mid \boldsymbol{X}) = \frac{P(\boldsymbol{X}, Y)}{P(\boldsymbol{X})}, \quad P(\boldsymbol{X}, Y) = N\left[\begin{bmatrix} \boldsymbol{\mu}_X \\ \mu_Y \end{bmatrix}; \begin{bmatrix} \Sigma_{\boldsymbol{XX}} & \Sigma_{\boldsymbol{X}Y} \\ \Sigma_{Y\boldsymbol{X}} & \Sigma_{YY} \end{bmatrix}\right] \qquad (96)$$

Explicitly, the conditional distribution can be expressed as [10-11]:

$$P(Y \mid \boldsymbol{X}) = N(\mu_{Y|\boldsymbol{X}}; \Sigma_{Y|\boldsymbol{X}}) = N(\beta_{Y,0} + \boldsymbol{\beta}_{Y,\boldsymbol{X}}; \Sigma_{Y|\boldsymbol{X}})$$

$$\mu_{Y|\boldsymbol{X}} = \mu_Y + \Sigma_{YX} \Sigma_{YY}^{-1} (\boldsymbol{X} - \boldsymbol{\mu}_X), \quad \Sigma_{Y|X} = \Sigma_{YY} - \Sigma_{YX} \Sigma_{\boldsymbol{XX}}^{-1} \Sigma_{\boldsymbol{X}Y} \qquad (97)$$

$$\beta_{Y,0} = \mu_Y - \Sigma_{YX} \Sigma_{YY}^{-1} \boldsymbol{\mu}_X, \quad \boldsymbol{\beta}_{Y,\boldsymbol{X}} = \Sigma_{YX} \Sigma_{YY}^{-1} \boldsymbol{X} \qquad (98)$$

Fig. 9 illustrates a three-generation linear Gaussian network, along with its means and variances [13]. The distribution for a continuous variable $Y$ with a discrete parent $Z$ and a continuous parent vector $\boldsymbol{X}$ is also a Gaussian distribution, conditional on parents' values.

$$P(A) = \mathcal{N}(\mu_A, \sigma_A^2)$$
$$P(B) = \mathcal{N}(\mu_B, \sigma_B^2)$$
$$P(C) = \mathcal{N}(\mu_C, \sigma_C^2)$$
$$P(D|A,B) = \mathcal{N}(\beta_{D,0} + \beta_{D,1}A + \beta_{D,2}B, \sigma_D^2)$$
$$P(E|C,D) = \mathcal{N}(\beta_{E,0} + \beta_{E,1}C + \beta_{E,2}D, \sigma_E^2)$$

**Fig. 9.** A Linear Gaussian Network [15]

$$P(y \mid \boldsymbol{x}, z_i) = P(\boldsymbol{x}, y, z_i) / P(\boldsymbol{x}, z_i) \equiv P^{(i)}(\boldsymbol{x}, y) / P(\boldsymbol{x}_i) = N(\mu_{y|\boldsymbol{x}}^{(i)}; \Sigma_{y|\boldsymbol{x}}^{(i)}) \qquad (99)$$

$$P^{(i)}(\boldsymbol{x}, y) = P^{(i)}(y \mid \boldsymbol{x}) \cdot P(\boldsymbol{x}) \qquad (100)$$

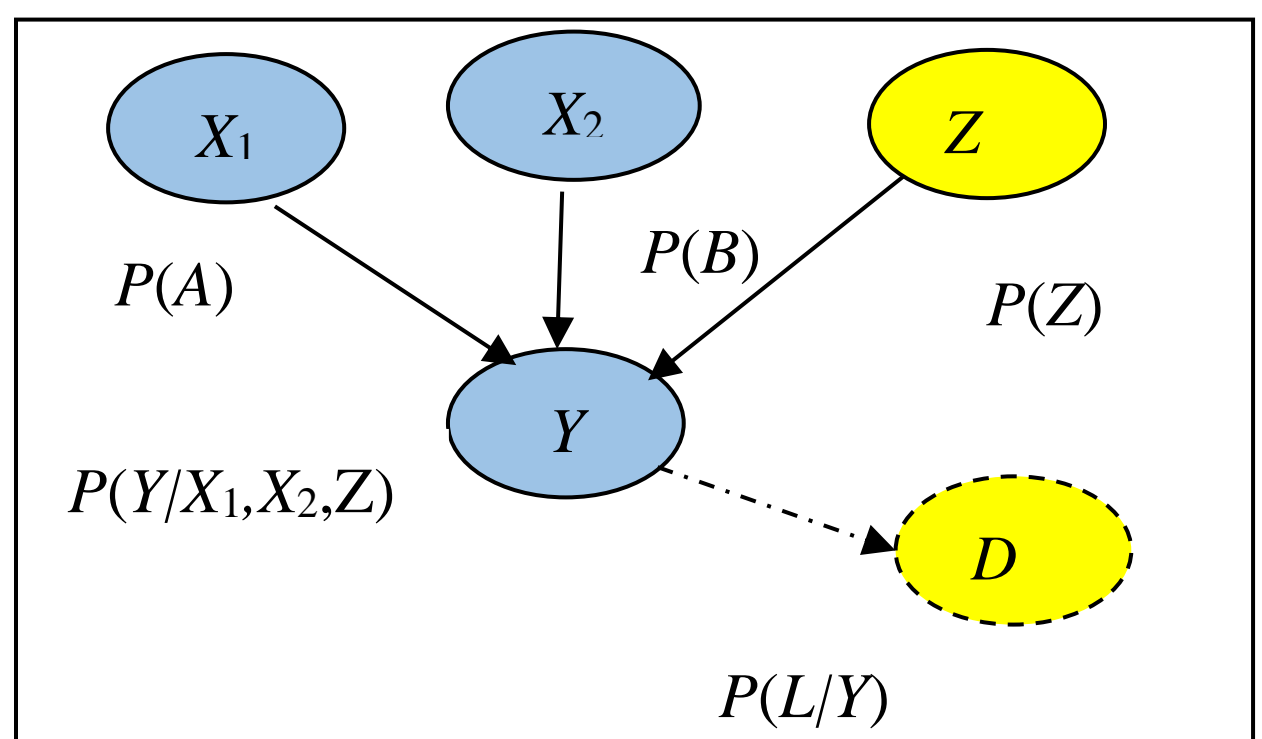

**Fig. 10:** A Mixed Bayesian Network
(Continuous blue; discrete: yellow; DD node: dashed)

This means that for each value $z_i$ of the discrete parent $Z$, there is a Gaussian distribution based on $z_i$ and its continuous parents. Fig. 10 shows such a network, where node $Y$ has two continuous parents, $X_1$ and $X_2$, and one discrete parent, $Z$.

Unlike discrete variables, for which software shows a list of possible states with their probabilities, as in Fig. 7, for a node with a continuous distribution, most software only shows the mean and the variance of the distribution. It passes only the mean value to its child (e.g., Fig. 3 in [14]).

**Discrete-Display (DD) Nodes:** However, if the network is used, e.g., to diagnose a patient's potential risk level of some disease, we propose to add a child node using the continuous distribution to calculate the probability for each level, building a table just like that for a discrete variable in the Student BN. We refer to this node as the Discrete-Display (DD) node because it discretizes and displays the continuous distribution of its parent node. The DD node's discrete probability table is derived solely from its parent's continuous probability distribution. Hence, a DD node is not a real random variable but a proxy of its parent. We use dashed arrows and shapes to represent such nodes, like node $D$ in Fig. 10, which don't contribute to the full joint probability distributions of the network. The DD nodes may be considered another discretization method for displaying continuous distributions [15, Sec. 4].

Moreover, to display user-specific results, continuous parent nodes, such as $X_1$ and $X_2$ in Fig. 10, should not simply pass the mean value down to their children; they would be better off using user-specific input, which can be implemented easily with GUI tools such as text fields, sliders, or plus-minus buttons.

Linear Gaussian regression is just one of many methods [16] for approximating random variables from collected data. The Gaussian distribution may not be a good choice if the data is highly asymmetric. We will not discuss the details. From now on, we assume we have already identified the required distribution functions (primary or conditional, discrete or continuous) for the Bayesian networks. They are at least integrable (either numerically or analytically). Eq. (99) provides the multi-conditional expectation value of $Y$ given $\{x_1, x_2, z_i\}$ for a Gaussian distribution. For a general distribution, by using Eq. (92) and (99), it can be expressed as follows, similar but not equivalent to Eq. (7) of [17]:

$$\begin{aligned} \overline{Y}_{Y|\boldsymbol{X},i} &\equiv E[Y \mid x_1, x_2, z_i] = P(\Omega \mid Y \mid x_1, x_2, z_i) = \int dy\, yP(y \mid x_1, x_2, z_i) \\ &\underset{(99)}{=} \int dy\, yP^{(i)}(y \mid x_1, x_2) \underset{(100)}{=} \int dy\, y\, P^{(i)}(x_1, x_2, y,) / P(x_1)P(x_2) \end{aligned} \tag{101}$$

In the last two steps, we have used Eqs. (99-100) and the fact that nodes $(X_1, X_2)$ are mutually independent. Similarly, the conditional variance is given by:

$$\Sigma_{Y|\boldsymbol{X},i} = \left[ \int dy\, y^2\, P^{(i)}(\boldsymbol{x}, y) / P(\boldsymbol{x}) \right] - \overline{Y}^2_{Y|\boldsymbol{X},i} \tag{102}$$

Suppose our DD node needs to show the possibilities for high and low risk. Assume that the range for low risk is $Y \in L_Y = [20, 100]$, high risk is $H_Y = (100, 200]$, and $P(y<20)+P(y>200)\ll 10^{-4}$. Using Eq. (93) and (99), the discretized conditional possibilities are given by the following multi-conditional probability expressions:

$$P(\text{low}_Y \mid \boldsymbol{x}, z_i) = P(L_Y / \boldsymbol{x}, \Omega_y, z_i) = \int_{20}^{100} dy P^{(i)}(\boldsymbol{x}, y) / P(\boldsymbol{x}) \propto \int_{20}^{100} dy P^{(i)}(\boldsymbol{x}, y) \tag{102}$$

$$P(\text{high}_Y \mid \boldsymbol{x}, z_i) = P(H_Y / \boldsymbol{x}, \Omega_y, z_i) = \int_{100}^{200} dy P^{(i)}(\boldsymbol{x}, y) / P(\boldsymbol{x}) \propto \int_{100}^{200} dy P^{(i)}(\boldsymbol{x}, y) \tag{103}$$

Hence, when we calculate the probability table of a DD node, it is more convenient to use its parent's joint probability density, and we can ignore the parent distribution's normalization coefficient by normalizing the DD results in the final stage:

$$P(\text{low}_Y \mid \boldsymbol{x}, z_i) \to P(\text{low}_Y \mid \boldsymbol{x}, z_i) / [P(\text{low}_Y \mid \boldsymbol{x}, z_i) + P(\text{high}_Y \mid \boldsymbol{x}, z_i)] \tag{104}$$

If necessary, one can also compute their multi-conditional expectations and variances:

$$\begin{aligned} E[L_Y \mid Y \mid \boldsymbol{x}, z_i] &\equiv P(L_Y \mid Y \mid \boldsymbol{x}, z_i) / P(L_Y \mid \boldsymbol{x}, z_i) \\ \Sigma_{L_Y|Y|\boldsymbol{x},i} &= E[L_Y \mid Y^2 \mid \boldsymbol{x}, z_i] - E[L_Y \mid Y \mid \boldsymbol{x}, z_i]^2 \end{aligned} \tag{105}$$

### 4.3. The Customized Healthcare BN with Mixed Random Variables

Let us build a static Bayesian network to predict a patient's blood sugar reading (*S*) in mg/dL, displaying the probabilities at its levels (*L*: low, normal, pre-diabetes, diabetes) and HDL cholesterol reading (*H*) in mg/dL, and displaying the probabilities at its levels (*D*: low, regular). We assume that blood sugar readings depend on daily hydrocarbon calories (C), average daily exercise time (*E*), and genetic inferences (*G*: no, weak, substantial). On the other hand, the HDL reading depends only on the amount of daily exercise (*E*). Altogether, we need five random variables (four continuous and one discrete) plus two DD nodes, as shown in Table 1 and Fig. 11.

| Symbol | Name | Type | Parent(s) | Unit/Levels |
|---|---|---|---|---|
| C | Calories | Continuous | No | Calorie |
| E | Exercises | Continuous | No | Minute |
| G | Genetics | Discrete | No | {0,1,2} |
| S | Sugar reading | Continuous | C, E, G | Mg/dL |
| L | Sugar level | DD node | S | {0,1,2,3} |
| H | HDL reading | Continuous | E | Mg/dL |
| D | HDL level | DD node | H | {0,1} |

**Table 1**: The random variables in the Healthcare BN

The Healthcare BN's DAG is shown in Fig. 11 below, which leads to the full joint distribution (DD nodes must be excluded) as follows:

$$P(C,G,E,S,H) = P(S \mid C,G,E) P(H \mid E) P(C) P(G) P(E) \tag{106}$$

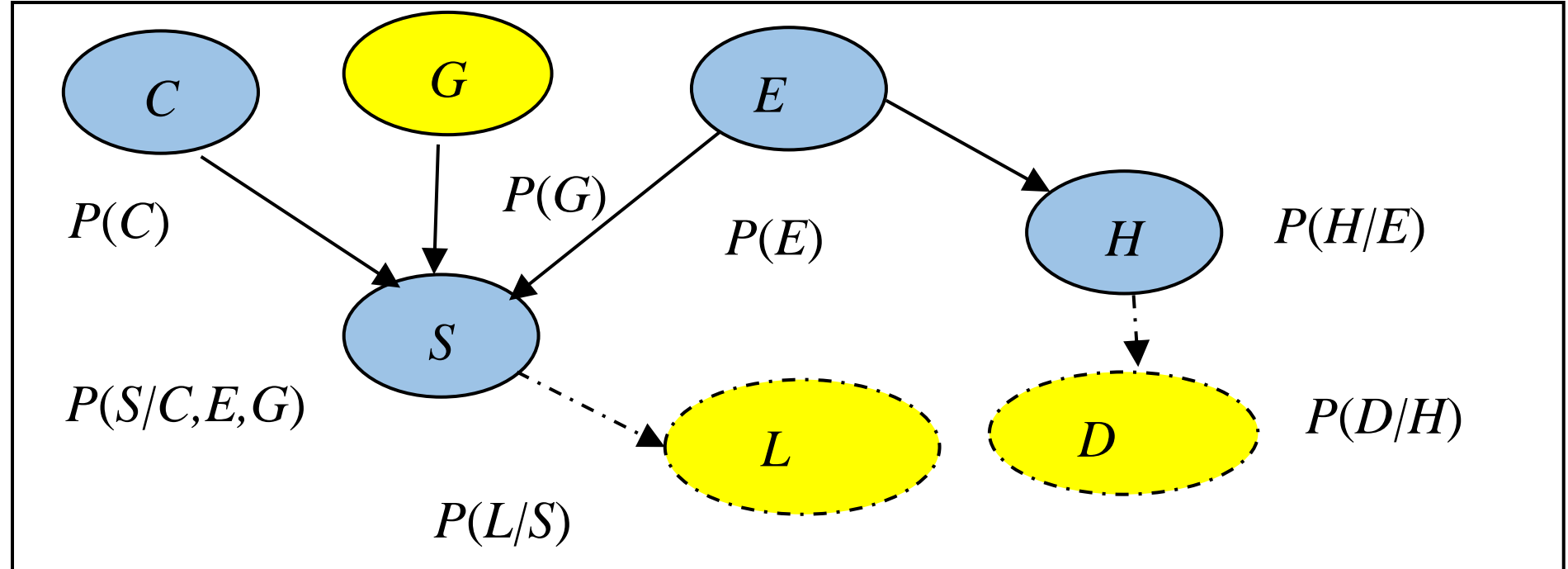

**Fig. 11**: The Customized Healthcare BN
(Continuous: blue; discrete: yellow; DD node: dashed

In the previous section, we detailed how to calculate various quantities: Eq. (94-102) for nodes *C*, *E*, *S*, and *H*, and Eq. (102-104) for the DD nodes *L* and *D*. How should these be displayed while the Healthcare BN is operating? Table 2 shows possible displays for four sample nodes.

| **C**: Calories in Carbs (Continuous) | |
|---|---|
| Mean | 1100 |
| Variance | 200.0 |
| User’s | 1300 |

| **G**: Genetic Inference (Discrete) | |
|---|---|
| No | 0.00 |
| Weak | 1.00 |
| Substantial | 0.00 |

| **S**: Sugar Reading (Continuous) | |
|---|---|
| Mean | 100 |
| Variance | 20.0 |

| **L**: Sugar Level (Discrete-display) | |
|---|---|
| Low | 0.04 |
| Normal | 0.10 |
| Pre-Diab. | 0.76 |
| Diabetes | 0.10 |

**Table 2**: The Displays of Nodes in the Healthcare BN

The continuous parent nodes (*C* and *E*) require user input to ensure user-specific predictions are made. These inputs can be entered in a text field initialized with the mean value and adjusted using GUI tools. The user’s selection determines the discrete parent *G*, similar to the Student BN. The Sugar Level DD node *L* predicts the user’s probabilities at each level.

Therefore, our customized Healthcare Bayesian Network (BN) uses user-specific data to generate tailored probability tables, similar to those produced by a BN with discrete variables. Such a customized BN, capable of handling continuous and mixed variables, can also be used in weather forecasting, predicting sensor equipment failure, healthcare, and other fields of interest.

## 5. Summary and Discussions

In this work, we extend Probability Bracket Notation (PBN) from single-variable and independent multivariable systems [1] to multivariable systems with dependencies and static Bayesian networks. Joint, marginal, and conditional distributions, as well as conditional expectations, are represented within a unified PBN framework. We show that the conditional probabilities and expectations underlying static Bayesian networks can be expressed concisely in a basis-independent form and computed using an operator-driven approach.

We formulate a general rule for inserting P-identities by introducing the concept of *d-separable chains* and demonstrate that, after a one-time preparation stage, the computational cost for a large *N*-node binary network can be reduced from $O(N{\cdot}2^N)$ based on the full joint probability table to $O(k{\cdot}2^k)$ for a d-separable inference chain with k intermediate nodes. PBN is particularly effective for d-separable chains in extended polytrees with pendant subnets (including pendant blobs, see Fig. 8). We also analyzed linear Gaussian networks and introduced a customized Healthcare Bayesian Network (Fig. 11) that supports both continuous and discrete random variables, enabling user-specific data input and displaying tailored predictions via Discrete-Display (DD) nodes that serve as proxies for their continuous parent nodes. Because expressions in PBN can be basis-independent, d-separable chains for inference or expectation may also be analytically useful for continuous random variables.

These results complement our earlier work on PBN in stochastic processes and path-integral formulations, as well as its applications to non-Hermitian Hamiltonians [1]. Together with our recent extension [7] to dynamic Bayesian networks [18], which can incorporate feedback mechanisms—a key limitation of static Bayesian networks [19], this work further establishes PBN as more than a notation: it is a structured computational framework with preprocessing and efficient inference phases. Beyond its pedagogical value in connecting Dirac-style notation with probability modeling, PBN provides a practical analytical tool for reasoning in complex probabilistic systems, with potential applications in Bayesian networks, machine learning (ML), and artificial intelligence (AI) [14,19,20].

## Abbreviations:

BN: Bayesian Network
CE: Conditional Expectation
CPD: Conditional Probability Distribution
DAG: Directed Acyclic Graph
DD node: Discrete-Display node
FJPT: Full Joint Probability Table
IPD: Intermediate Probability Distribution
JPD: Joint Probability Distribution
MPD: Marginal Probability Distribution
P-basis: Probability-basis (P-bra, P-ket, P-bracket, P-identity)
PBN: Probability Bracket Notation,
PD: Probability Distribution, Probability Density
PGM: Probabilistic Graphic Model

## Appendix A: The Two-Phase Chain Inference Process and Its Efficiency

Here, in this appendix, we discuss the two-phase procedure for computing inference for d-separable chains in PBN and its efficiency.

**A.1: The Preprocess and the Computation of Chain Inference in PBN**

The primary advantage of PBN is its ability to efficiently handle *d-separable inference chains* in large-scale $N$-node extended polytrees (e.g., $N$~100). Instead of computing a complex conditional probability, e.g., $P(A|F)$, directly from a full joint distribution table (FJPT), PBN allows for the strategic insertion of P-identities for intermediate nodes $I_C$, $I_D$ ..., effectively calculating P $(A|I_C I_D... |F)$ for a d-separable chain on the backbone polytree. This breaks a large problem into a sequence of simpler, pre-computed operations. To leverage this for scalable inference, a practical implementation should first pre-calculate the following local tables for the network and store the local MPTs and CPTs:

1. Local parent-child JPTs: $P\ (X_C, X_P)$, optional for storage.
2. Local marginal probability tables (MPTs) for each node: $P(X)$. (A.1)
3. Local parent-child and child-parent CPT: $P\ (X_C|\ X_P)$, and $P(X_P/X_C)$.

The tables can be derived using Bayes' formula and the definition of an MPT starting from the root parents. For example, in the Student BN, the pre-calculation process can start at the first CPT in the right of Eq. (60) to produce and store the local probability MPT and CPT tables in the following order (there are two inference chains):

$$P(d), P(i), P(g), P(g/d), P(d/g), P(g|\ i), P(i|g), P(l); P(s), P(s|i), P(i|s) \quad \text{(A.2)}$$

The total number of operations required to populate local tables depends on the structure of the BN. The operation count (#c) for a next-generation node depends on the number of its outcomes, the number of its parents, and the independence of its parents. In a large $N$-node binary network, it can be shown that, for a node with a single parent, e.g., P(c|a)[3], #c ~ 8; for a node depending on 2 parents, e.g., P(c|x,y), #c ~ 52 if X and Y are independent[4]; otherwise, it is a pendant blob at C. One should start from the top parent in the blob to calculate the #c of the subnet, and each node adds #c ~ O(10). For example, if X and Y are directly linked, X→Y (as in Fig 8), #c ~ 8 for Y. Note that any inference chain that starts or ends in a pendant blob (except node C) may not be d-separable, and a pendant blob does not affect the d-separable chains passing through node C on the polytree backbone (see §3.4.1 and Fig. 8); we can ignore them if we are only interested in the d-separable chains on the backbone polytree. The analysis applies to nodes depending on more than 2 parents: if a node depends on $M$ independent parents, #c ~ $M \cdot 2^{M+2}$, in general $M \ll N$. Therefore, the upper limit of preprocess operations can be set to #c ~ $N \cdot 2^N$ for a network with a polytree backbone.

Altogether, preparing all local tables for $N$ nodes requires $O(10 \cdot N)$ to $O(N \cdot 2^N)$ operations. For example, the preparation for the Student BN involves about 100 operations. After these tables are stored as data objects associated with the BN, an inference task reduces to identifying the d-separable sequence of intermediate nodes between two target nodes on the backbone polytree. Such an algorithm is straightforward to implement in languages such as Python or Java and is well-suited for AI-assisted symbolic reasoning. Once the d-separable chain between query nodes is determined (e.g., A→C←D→E), the unknown conditional probability $P(A|E)$ is computed by sequential summation:

$$P(a\,|\,e) = P(a\,|\,I_C I_D\,|\,e) = \sum_{c,d} P(a\,|\,c)P(c,d)P(d\,|\,e) \rightarrow \#c = 4\cdot 2\cdot 4 = 4(k+1)\cdot 2^k \sim k2^{k+2} \quad \text{(A.3)}$$

[3] The four 2-node joint, P(a,c) = P(c|a)P(a) → #c = 2^2 = 4; the two single-node marginal: P(c) → #c = 2 • 2 = 4.

[4] The eight 3-node joint, P(a,b,c) = P(c|a,b)P(a)P(b) → #c = 2^3 • 2 = 16; the two 2-node marginal joints, P(a,c) and P(b,c) → #c = 2 • 2^2 • 2 =16; the two single-node marginal, P(c) → #c = 2 • 2 = 4; the four CPTs: P(a|c), P(c|a), P(b|c), P(c|b) → #c = 4 • 2^2=16.

Consider a BN with $N$ binary nodes and an inference chain involving $k$ intermediate nodes. The CP has 4 matrix elements; for each of the elements, the total number of summations is $2^k$, and there are $k+1$ multiplications, leading to approximately $4\cdot(k+1)\cdot 2^k \sim k\cdot 2^{k+2}$ operations. Altogether, PBN needs $O(10\cdot N)$ to $O(N\cdot 2^N)$ operations to prepare, and $O(k\cdot 2^k)$ to calculate the CP for a given inference chain.

**A.2. The Comparison of Efficiency in Chain Inference**

In contrast, a brute-force approach based on the FJPT (which is a product of $N$ given marginal or conditional distributions), although no need for preparation, would require $N\cdot 2^{N-2}$ operations to find the CPT of the reasoning chain.

$$P(a\,|\,e) = \frac{P(a,e)}{P(e)} = \sum_{\text{all but } a,e} \frac{1}{P(e)}\{\text{FJPT}(a,b,c,d...)\},\ P(e) = \Sigma_a P(a,e),\ C_\# \sim N2^N \quad \text{(A.4)}$$

Therefore, after preparation, it is the PBN's $O(k\cdot 2^k)$ vs. FJPT's $O(N\cdot 2^N)$. The ratio is:

$$(k/N)\cdot(2^k/2^N) \sim (k/N)\, 2^{k-N} \quad \text{(A.5)}$$

For $N \sim 100$ and $k \sim 10$, the number of calculations is reduced by a factor of $2^{90} \sim 10^{27}$, which is nearly exponential. This exponential gap highlights why PBN's calculation, based on local d-separable reasoning chains, is vastly more efficient and scalable than using FJPT in large Bayesian networks. The computational advantage of PBN arises from a *two-phase strategy*: an initial preprocessing stage that constructs local probability tables, followed by efficient query evaluation using P-identities. Although the preprocessing cost may be comparable to a single brute-force inference computation, it is incurred only once, whereas subsequent inference queries are significantly accelerated.

***An Illustrative Example of Inference via P-Identities***: Suppose the inference of interest is $P(D|L)$ in the Student BN, whose structure indicates that the influence of $L$ on $D$ propagates through $G$, which d-separates $L$ from $D$: $L\rightarrow G\rightarrow D$. In PBN, one can insert the corresponding P-identity $I_G$, make explicit the inference chain encoded in the BN, and avoid any reference to the full joint distribution (for the summation range, see Fig 7).

$$\text{PBN: } P(d\,|\,l) = P(d\,|\,I_G\,|\,l) = \sum_{g=1}^{3} P(d\,|\,g)P(g\,|\,l) \rightarrow \#c = 4\times 3 \approx 10 \quad \text{(A.6)}$$

$$\text{FJPT: } P(d\,|\,l) = \frac{P(d,l)}{P(l)} = \frac{\sum_{i,g,s} P(i)P(d)P(g\,|\,i,d)P(g\,|\,l)P(s\,|\,i)}{\sum_d P(d,l)} \rightarrow \#c = 4\times 12 + 4 \approx 50 \quad \text{(A.7)}$$

Therefore, even in this relatively small network, the PBN-based computation reduces the operation count by a factor of several times. This advantage becomes significantly more noticeable in large-scale networks, where the complexity of brute-force marginalization grows exponentially with the number of variables.

***Conditional Expectation via P-Identities***: If the nodes have numerical outcomes, we can compute conditional expectations (e.g., predicted precipitation on specific days in a complex weather model). The same stratagem applies. For example, we have all numerical outcomes for the Student BN in Eq. (58) and want to compute the expectation of S given L. The inference chain is S←I →G →L, which is d-separable. Following Eq. (83-85), we have:

$$E[S\,|\,l] \equiv P(\Omega\,|\,S\,|\,l) = P(\Omega\,|\,SI_S I_I I_G\,|\,l) = \sum_{s,i,g} sP(s\,|\,i)P(i\,|\,g)P(g\,|\,l) \tag{A.8}$$

The *bridged expectations*, analogous to Eq. (36), can be calculated in a similar way, e.g.:

$$E[i\,|\,G\,|\,l] \equiv \frac{P(i\,|\,G\,|\,l)}{P(i\,|\,l)} = \frac{P(i\,|\,I_I G I_G\,|\,l)}{P(i\,|\,I_G\,|\,l)} = \frac{\sum_g P(i\,|\,g)gP(g\,|\,l)}{\sum_g P(i\,|\,g)P(g\,|\,l)} \tag{A.9}$$

These cases show that PBN can greatly reduce the computation time while explicitly revealing the chain of inferences for conditional probabilities and expectations in Bayesian networks.

Moreover, assume that the nodes in the Student BN are continuous random variables, then Eqs. (A.8-9) are still analytically valid if summations are replaced by integrals, and conditional probability tables are replaced by corresponding joint and marginal probability densities, as was done in Eqs. (92-93).

***Relation with Variable Elimination****:* It is worth noting that the PBN-based inference chain processing is related to the *variable elimination* method commonly used in Bayesian networks (Chap. 9, [3]). In both approaches, exact inference is achieved by selecting appropriate intermediate variables, thereby avoiding the need to use the full joint probability distribution. PBN is particularly effective for d-separable chains with pendant subnets on *Extended Polytrees*, as defined in §3.4.2 (an example is shown in Fig. 8). PBN provides a compact, operator-based process for computing chain inference or expectation, where inserting identities makes the sequence of selections explicit and algebraically transparent.